\documentclass[runningheads]{llncs}
\usepackage{graphicx}
\usepackage{amsmath,amssymb} 
\usepackage{color}\usepackage[width=122mm,left=12mm,paperwidth=146mm,height=193mm,top=12mm,paperheight=217mm]{geometry}
\usepackage{xspace, subfigure}

\newcommand{\mat}[1]{\mathbf{#1}}
\renewcommand{\vec}[1]{\mathbf{#1}}
\newcommand{\Silhouette}{{\cal I}}

\begin{document}
\pagestyle{headings}
\mainmatter
\title{A model-based approach to recovering the structure of a plant from images} 

\titlerunning{A model-based approach to recovering the structure of a plant from images}

\authorrunning{B. Ward et al.}

\author{Ben Ward\inst{1} \and John Bastian\inst{1} \and Anton van den Hengel\inst{1}
\and Daniel Pooley\inst{1} \and Rajendra Bari\inst{2} \and Bettina Berger\inst{3} \and Mark Tester\inst{4}
}

\institute{School of Computer Science, The University of Adelaide, Adelaide, Australia\\
\email{\{ben.ward, john.bastian, anton.vandenhengel, daniel.pooley\}@adelaide.edu.au}
\and Bayer CropScience, Ghent, Belgium\\
\email{rajendra.bari@bayer.com}
\and The Plant Accelerator, The University of Adelaide, Adelaide, Australia\\
\email{bettina.berger@adelaide.edu.au}
\and Center for Desert Agriculture, King Abdullah University of Science and Technology, Thuwal, Saudi Arabia\\
\email{mark.tester@kaust.edu.sa}}
\maketitle

\begin{abstract}
We present a method for recovering the structure of a plant directly from a small set of widely-spaced images. Structure recovery is more complex than shape estimation, but the resulting structure estimate is more closely related to phenotype than is a 3D geometric model. The method we propose is applicable to a wide variety of plants, but is demonstrated on wheat.  Wheat is made up of thin elements with few identifiable features, making it difficult to analyse using standard feature matching techniques. Our method instead analyses the structure of plants using only their silhouettes. We employ a generate-and-test method, using a database of manually modelled leaves and a model for their composition to synthesise plausible plant structures which are evaluated against the images.  The method is capable of efficiently recovering accurate estimates of plant structure in a wide variety of imaging scenarios, with no manual intervention.

\end{abstract}

\section{Introduction}

Computer vision techniques can provide fast, accurate, automated, and noninvasive measurements of phenotypic properties of plants. Measurements of properties such as volume, leaf length, and leaf angle can be used to evaluate the effect on plants of variation in environmental conditions or genetic properties \cite{FuncStruc}. Obtaining these measurements from image data can be difficult. Plants typically have properties such as uniform colour, specular surfaces, and thin regions which present challenges for typical reconstruction techniques.

Existing methods have focused largely on the problem of recovering the shape of the plant independent its structure. Structure, here, is intended to encompass the various parts of a plant and the relationships between them, as opposed to purely geometric shape information contained in a representation of the whole plant such as a 3D volume or point cloud.  The structure may be represented in terms of anatomical aspects of the plant, but may equally be described in terms of more basic elements.  Importantly, structure allows the application of prior knowledge about the grammar of particular types of plants.

There are two primary advantages of considering structure rather than shape.  The first is that structure is much more closely related to plant anatomy, and therefore a much better indication of phenotype.  The second advantage is that the structural properties of a plant provide a strong indication of the likelihood of a particular shape, which is a valuable cue when trying to select from among multiple feasible shapes.  For plants with potentially complex structures, such as wheat, there may be many possible plant shape hypotheses which are supported by an image set, whereas prior knowledge of plant anatomy may indicate that only one structure is feasible.  This means that structure recovery is possible when shape estimation alone would be ambiguous, or equivalently, that fewer cameras are required to estimate structure than shape.  A related advantage is that even if more than one shape is supported by the image set, these shapes often have closely related structures, so although the images are ambiguous (in terms of shape) they may still support an estimate of structure, and thus a phenotypic interpretation.

Structure, for the purposes of the method we propose here, includes information about the identity, length, and curvature of each leaf in the plant, and the relationships between leaves. In this method each leaf is represented by a 3D curve tracing the central axis of the leaf from its tip to the base of the plant. The combination of multiple leaf models gives a model of a complete plant. This estimate of structure implies a particular 3D shape of the plant, which may be used to estimate which pixels belong to each plant element. Estimating structure thus enables the length of leaf 4 on day 10 to be measured, for example, and post-processing would allow an estimate of the width or the length of senescence.

\begin{figure}[!tb]
\centerline{
\includegraphics[width=0.18\columnwidth]{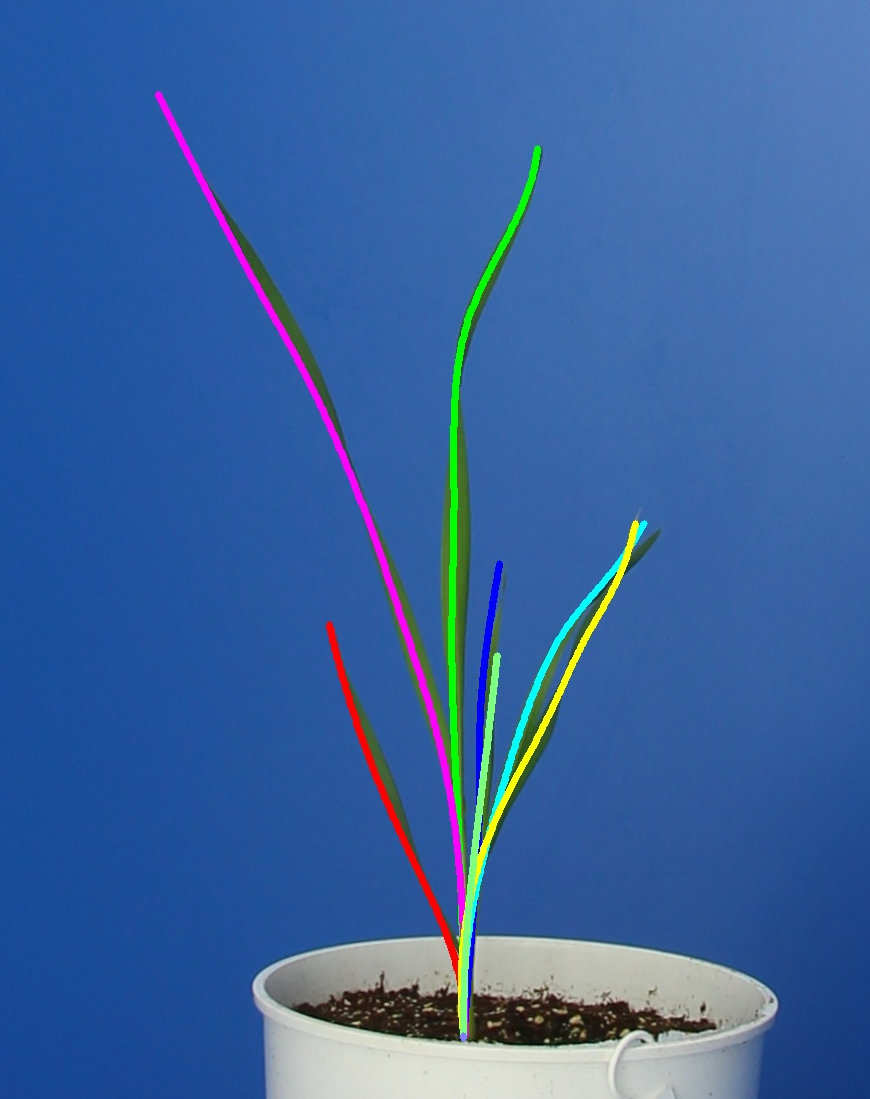}
\includegraphics[width=0.18\columnwidth]{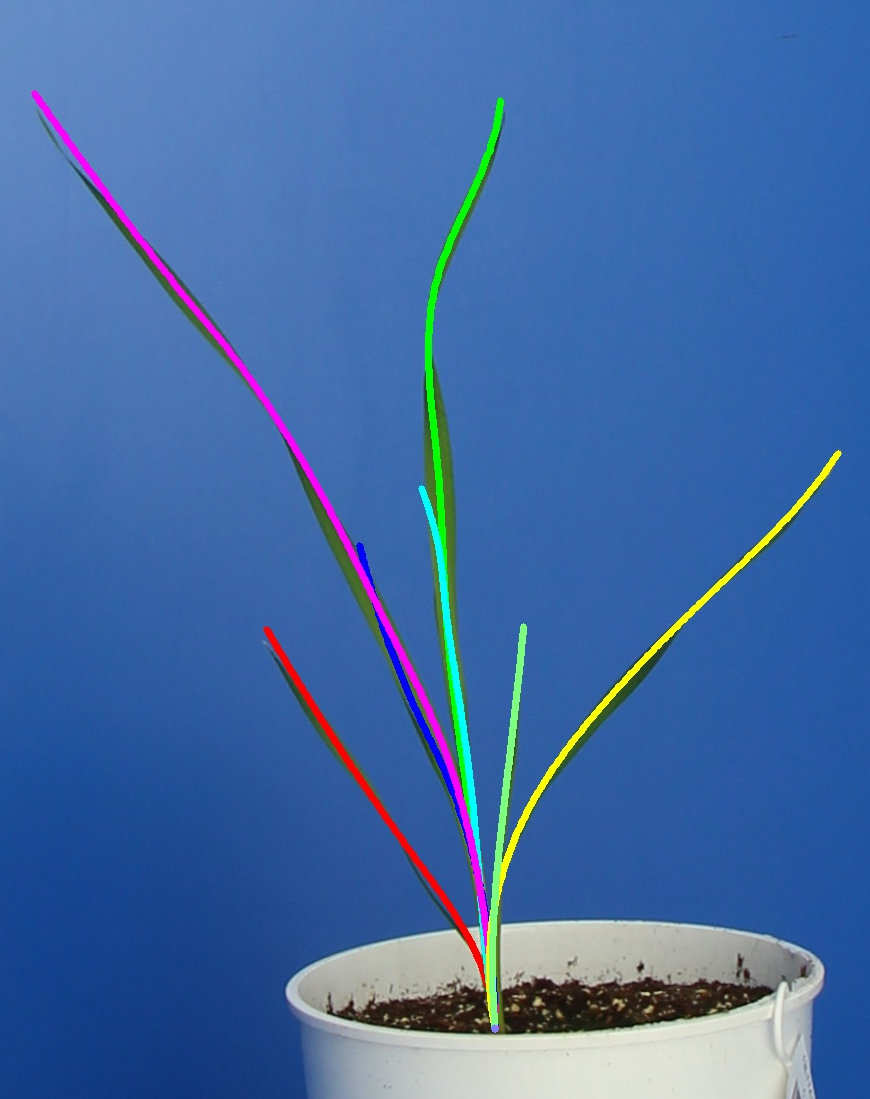}
\includegraphics[width=0.18\columnwidth]{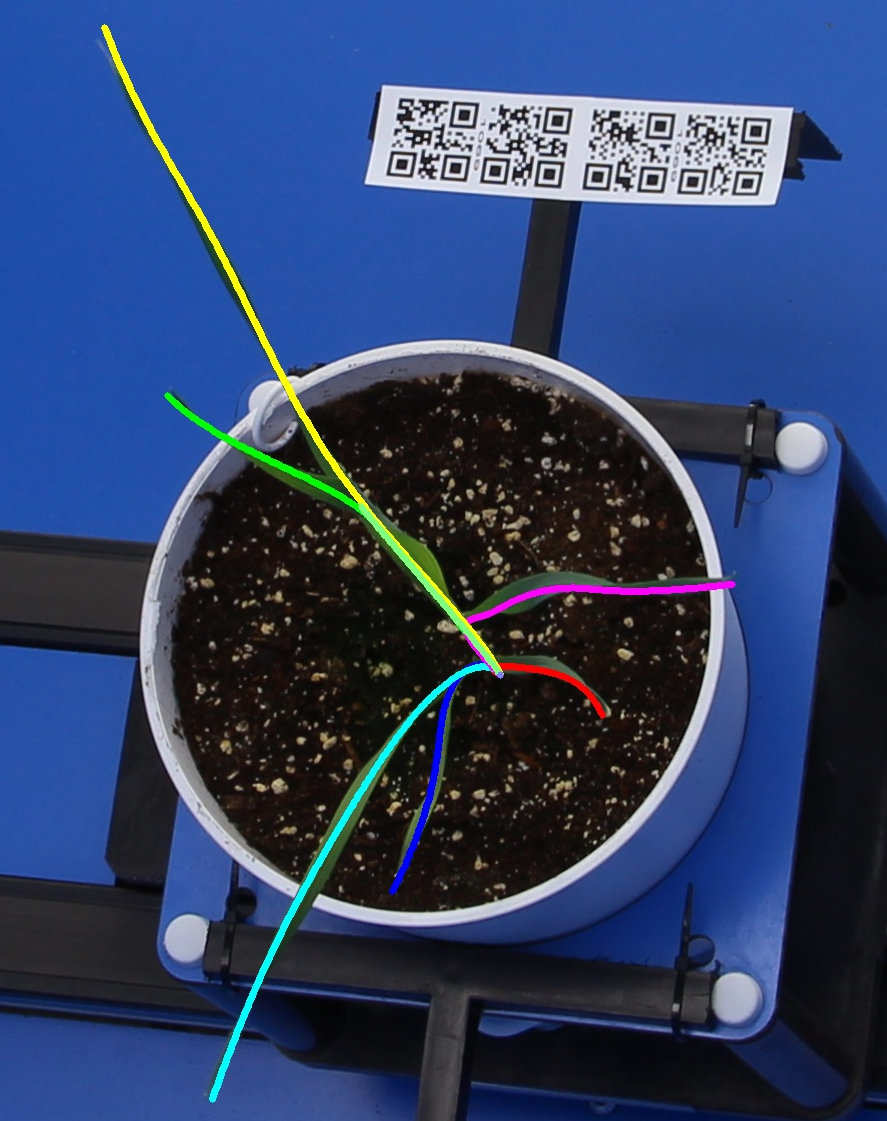}
\includegraphics[width=0.18\columnwidth]{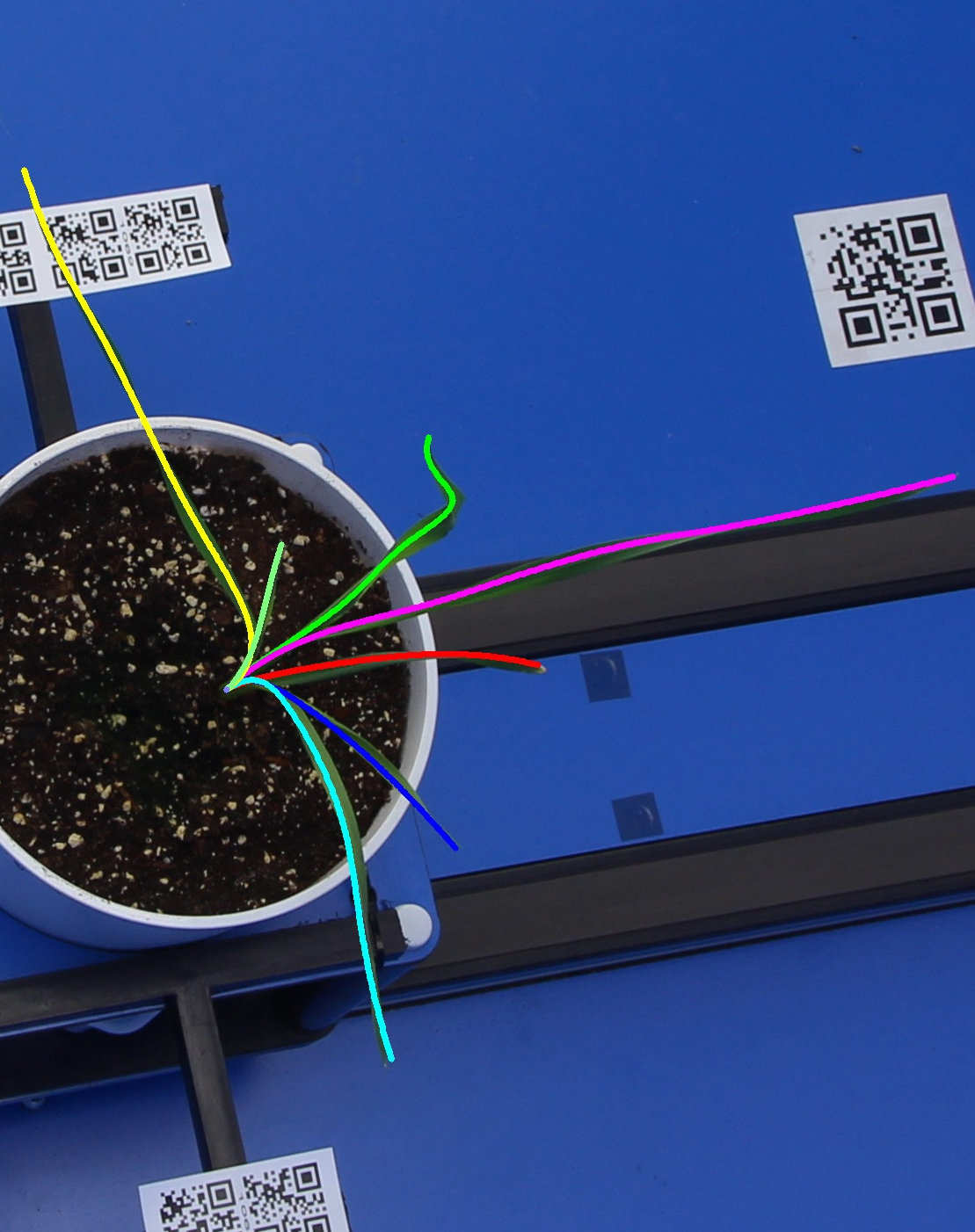}
\includegraphics[width=0.18\columnwidth]{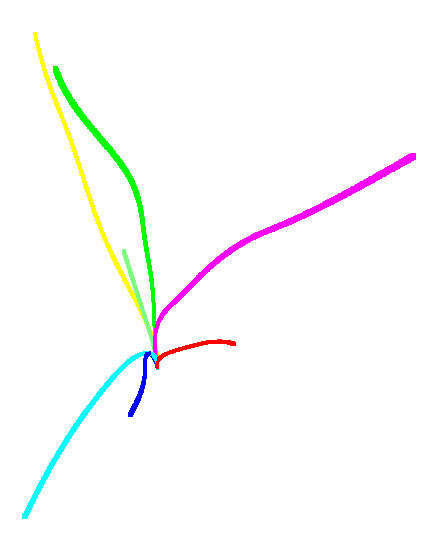}
}
\caption{A reconstructed plant model projected into the original images, and another view of the 3D model}
\label{fig:recon1}
\end{figure}

The method we describe is capable of estimating the structure of a plant made up of thin elements from a small set of images taken from widely-spaced viewpoints. Because the properties of these plants make reconstruction difficult using standard feature matching techniques, we reconstruct the plants using only their silhouettes. We employ a generate-and-test method, generating possible plant structures which are evaluated against the images. The generation process makes use of a database of leaf models, providing prior information on plausible leaf curves, which we use to restrict the generated models to plausible plant structures. The space of possible generated models is therefore significantly smaller than if we were to generate models by naively sampling 3D curves, allowing for a more efficient reconstruction process. Likely leaf tip locations are also detected, and used to further constrain the space of possible models. Figure~\ref{fig:recon1} shows a 3D plant model estimated with this method projected into the original image set.

\section{Related Work}

A range of techniques currently exist for automated extraction of phenotypic properties from image or depth data. Highly detailed and accurate point-cloud reconstructions can be obtained with the use of technology such as laser scanners \cite{SurfaceBased} or structured light \cite{GrowingPlants,Chlorophyll}. However, this technology can be prohibitively expensive or infeasible to incorporate into existing systems, may not provide sufficient resolution for recovering thin structures, and can be difficult to apply when plant size varies greatly. Reconstruction from images can provide a lower cost and a more practical solution. Methods based on identifying plant pixels can be used to estimate volume without recovering 3D structure \cite{HighThrough1,ShootBiomass}. Image based approaches for recovering 3D reconstructions employ techniques such as dynamic programming \cite{StereoVis} and simulated annealing \cite{SimAnneal} to overcome the difficulty of identifying corresponding points between frames. Reconstruction based on matching line features can provide robustness to appearance variation in different views \cite{LinFeat,CornRecon,CurveBased}. Complex plant structures with overlapping leaves mean a large number of views of the plant are usually required for a complete reconstruction. Techniques for obtaining a dense set of views of a plant include the use of mirrors \cite{HighThrough} or cameras mounted on robotic arms \cite{ToFData}. Mechanisms for turning the plants \cite{HighThrough1} can be used to generate a range of views, but can cause leaf movement which leads to additional difficulties for reconstruction. For methods which recover a point cloud or volumetric description of a plant, additional processing such as applying skeletonisation operations to a point cloud is required to recover a structural description \cite{LaserScan}. Interactive methods avoid some of the difficulty of fully automated techniques \cite{PlantModelling,MorphTraits} but significantly increase the time and manual effort required for reconstruction.

\section{Method}

We aim to recover an estimate of the length, curvature, and identity of each leaf of a grass plant, in this case wheat, from a set of images.
The image set may be small (the results in this paper were obtained from four images), and captured with widely-spaced cameras. Widely-spaced views, and the thin components and relatively uniform colour of these plants, make accurate reconstruction infeasible using standard feature matching techniques. Such techniques would also not provide data on the structure of the plant in occluded regions.

\begin{figure}[!tb]
\centerline{
\includegraphics[width=0.25\columnwidth]{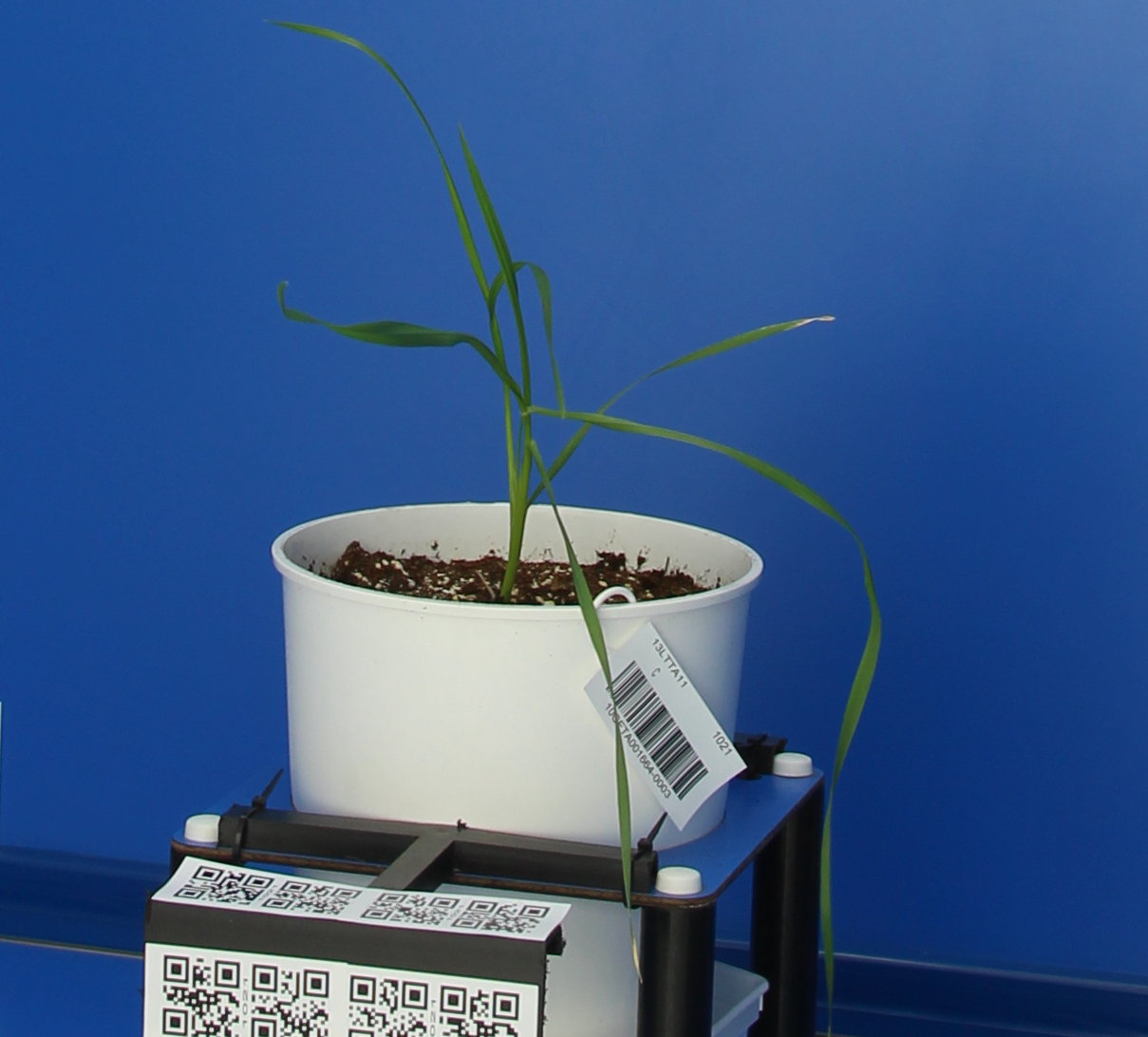}
\includegraphics[width=0.25\columnwidth]{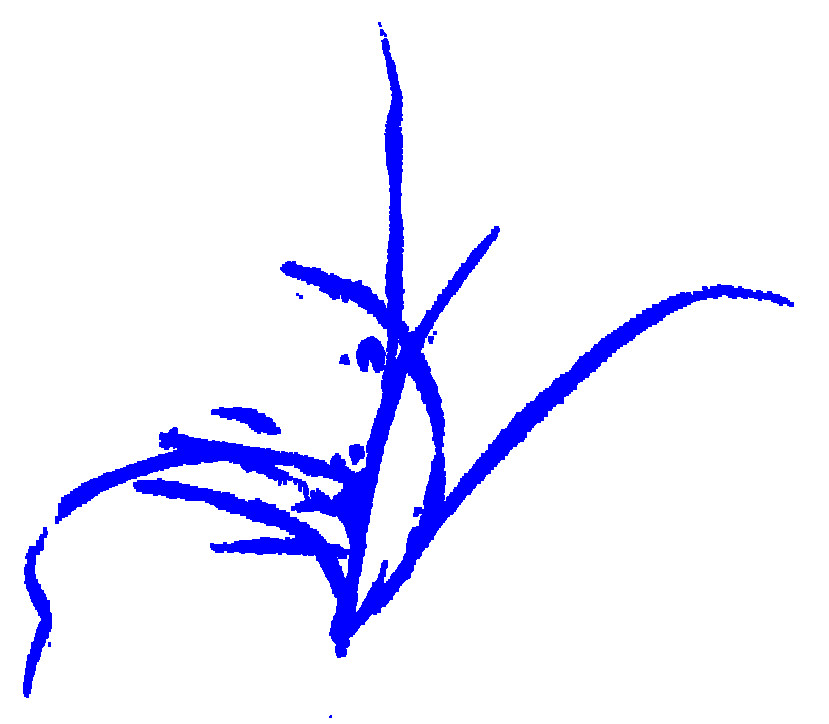}
\includegraphics[width=0.25\columnwidth]{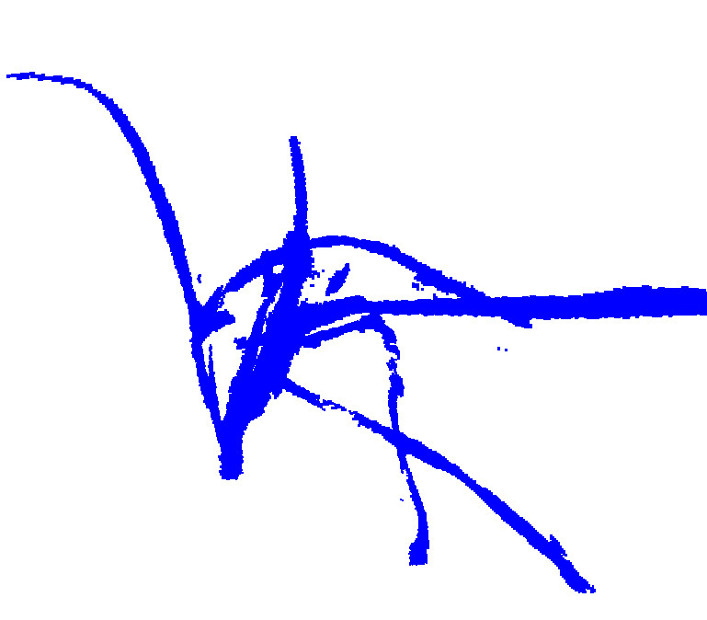}
\includegraphics[width=0.25\columnwidth]{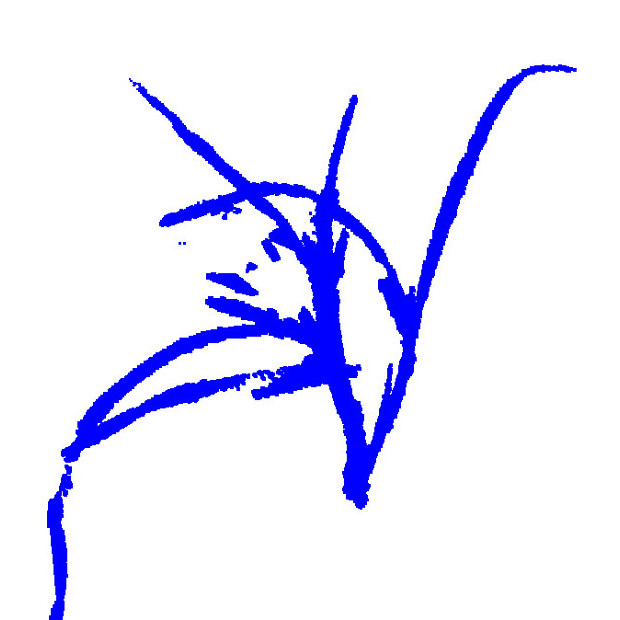}
}
\caption{A visual hull reconstruction illustrating the spurious shapes beyond the true plant reconstruction which are inherent to the visual hull.}
\label{fig:vishull}
\end{figure}

Given that attempting to match the appearance of individual points on leaves is infeasible, we instead analyse the silhouette of the plant in each view. Using standard silhouette-based reconstruction methods \cite{VisHull} could leave the 3D structure ambiguous when only a small number of views is available. Figure \ref{fig:vishull} shows three views of the visual hull generated from the four silhouettes for the plant on the left. Due to self occlusion and the limited set of views, this visual hull reconstruction includes leaf-like regions which do not correspond to actual leaves of the plant.

Instead of directly recovering the 3D shape of the plant from the silhouettes, we use a generate-and-test method to recover the 3D structure, generating plausible 3D plant models and evaluating them against the image set. A process of using prior knowledge to generate plausible structures which are evaluated against the data is employed for tree and plant reconstruction by methods such as \cite{AnalBySyn,Unfoliaged,StatTrees}.

This method allows us to use prior information about the plants being reconstructed to aid in determining the structure in regions where that structure would be ambiguous given only the image data. We make use of a database of manually modelled leaves. The reconstruction process generates 3D plant models by finding leaf models in the database which closely match the current image set, then refining these individual leaf models, and selecting an optimal combination of leaf models to model the complete plant.

\subsection{Input Data}

The input to our process is a set of images of a plant. The method is suitable for use with any number and placement of cameras, provided two views are available for each leaf. For results in this paper, we used four images captured by cameras covering $360^{\circ}$ around the plant. These images were captured with a set of consumer-grade DSLR cameras. The method requires calibrated cameras with known scale. We also require the approximate location of the centre of the pot, and a vector giving the vertical orientation of the pot. To obtain the necessary calibration information with minimal manual intervention, we make use of a calibration object providing features on multiple planes in each view. A 3D model giving the approximate structure of the pot and pot holder is also used to estimate occlusion. We require that the leaves are static while images are being captured, and that the leaves do not move between images being captured.

The structure recovery process estimates a silhouette of the plant for each frame.
Depending on the background of the scene, a colour histogram thresholding method (as applied, for example, in \cite{HighThrough1}) may be sufficient. Due to the variation in colour and texture of the plants and background in the image sets we are using, a pixel classifier using a Support Vector Machine trained on manually labelled images was applied for the mask generation.

\subsection{Calibration}

Camera calibration is achieved using a calibration object that displays known patterns to a variety of viewpoints. Rather than using independent planes as proposed in \cite{sturmcalib} and \cite{zhangcalib}, a single rigid object is favourable here as it does not require that fixed cameras view planes in common. Such an object can also be placed within an automated greenhouse system so that calibration can be periodically performed or verified.  The shape of the object is recorded in a file such as may be sent to one of the many acrylic laser cutting services so it can be rapidly constructed anywhere in the world.

QR codes are used as calibration patterns as they are rich in features and can be uniquely identified. The patterns are printed onto durable adhesive labels for robustness against humidity and temperature. The adhesive labels are placed  onto the object manually, resulting in some ambiguity in their true locations. Rather than rely on a large number of manual measurements, adhesive label placements are described by calibration object parameters which are estimated as part of the calibration process.

\begin{figure}[t]
\centering\leavevmode
\begin{minipage}[t]{0.35\linewidth}\centering\leavevmode
\includegraphics[width=0.9\linewidth]{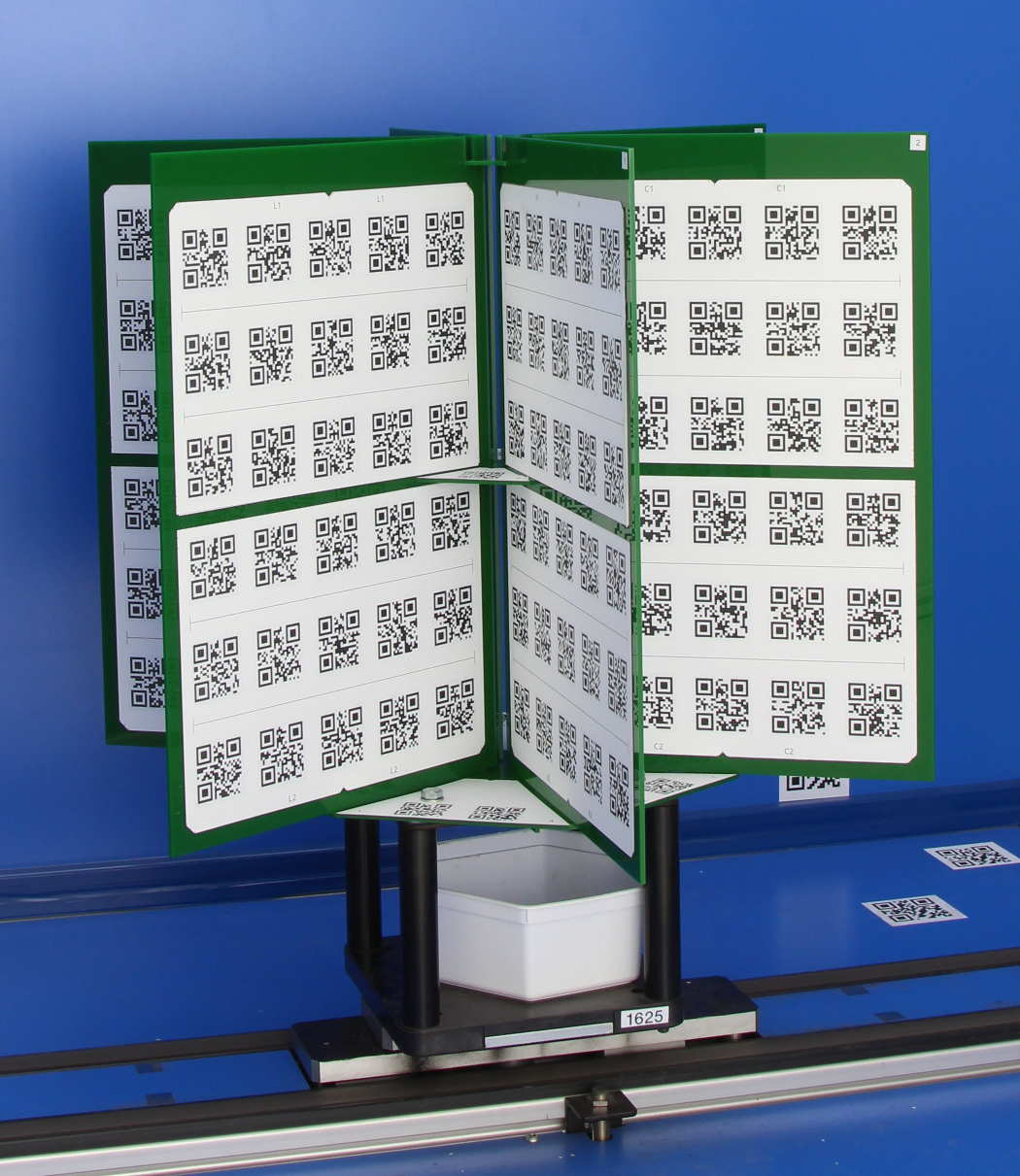}
\caption{The calibration object}
\label{fig:calib}
\end{minipage}\hspace*{0.03\linewidth}
\begin{minipage}[t]{0.6\linewidth}\centering\leavevmode
\includegraphics[width=0.48\linewidth]{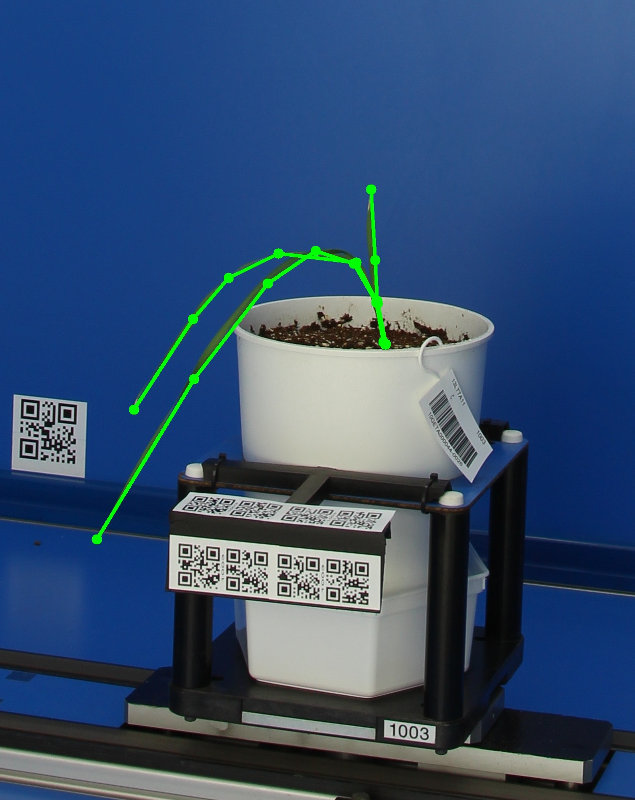}
\includegraphics[width=0.48\linewidth]{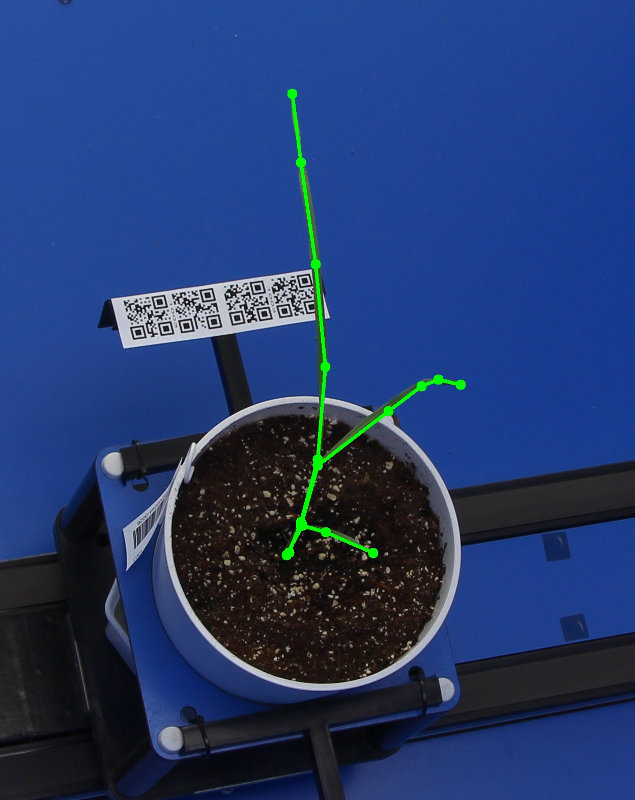}
\caption{A plant model manually constructed from two views}
\label{fig:manual}
\end{minipage}
\end{figure}

Initial camera poses and intrinsic parameters are estimated assuming ideal (known) placements of calibration patterns. Subsequently, both camera parameters and calibration object parameters are refined so as to minimise the sum of squared reprojection distances and error terms based on prior estimates of the calibration object parameters. Figure~\ref{fig:calib} shows the calibration object.

\subsection{Database Building}

The goal of the method is to generate a plausible plant model given the silhouette in each view. To generate candidate models representing plausible plant structures, we use a database of pre-defined leaf models. These models are manually constructed using an interactive tool. Plants are modelled by specifying a series of 3D point locations tracing the axis of each leaf. To specify a point location, the user first selects a point on a leaf in one view of the plant, then selects the corresponding point in a second view. The selected point in the second view is constrained to lie on the corresponding epipolar line. The database currently contains models for 480 leaves, modelled from 230 plants. Each leaf is modelled with an average of $8$ points. Figure~\ref{fig:manual} shows two views of a manually modelled plant. To increase the density of the database, additional leaf models are generated by transforming modelled leaves to stretch their shape in multiple directions within a small distance range. This generates $100$ models for each modelled leaf.

\subsection{Skeleton Extraction}

\begin{figure}[t]\leavevmode\centering
\begin{minipage}[t]{0.52\linewidth}\leavevmode\centering
\subfigure[Input]{\includegraphics[width=0.31\columnwidth]{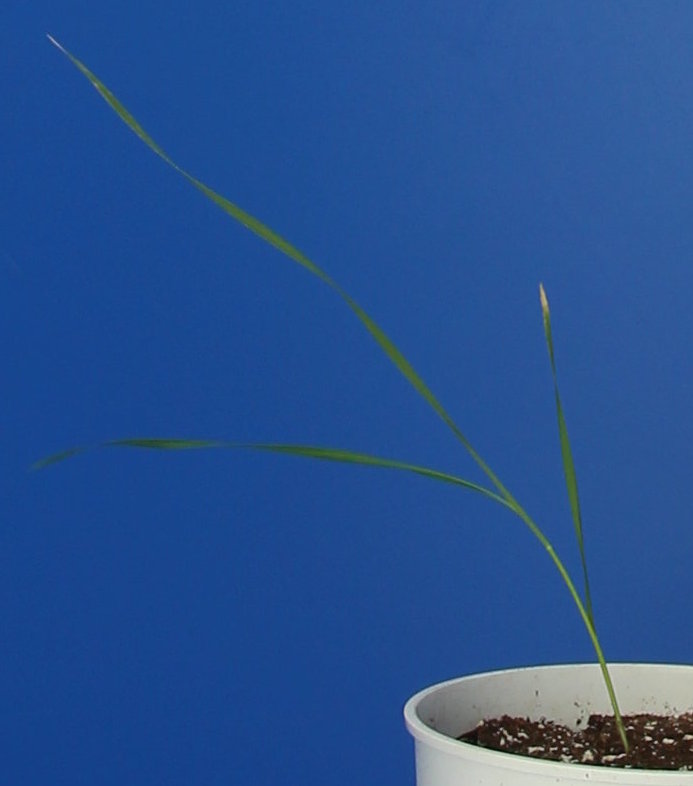}}
\subfigure[Classified]{\includegraphics[width=0.31\columnwidth]{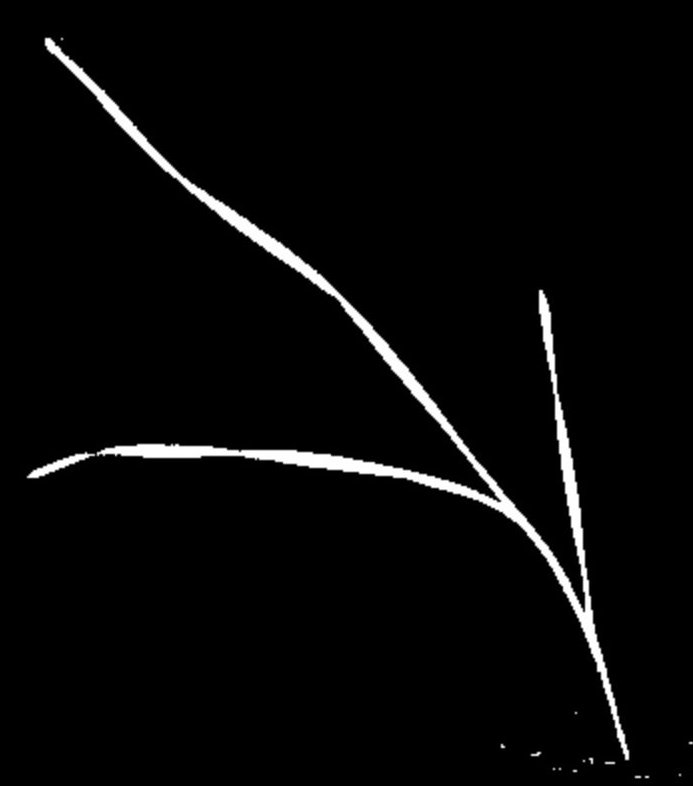}}
\subfigure[Skeleton]{\includegraphics[width=0.31\columnwidth]{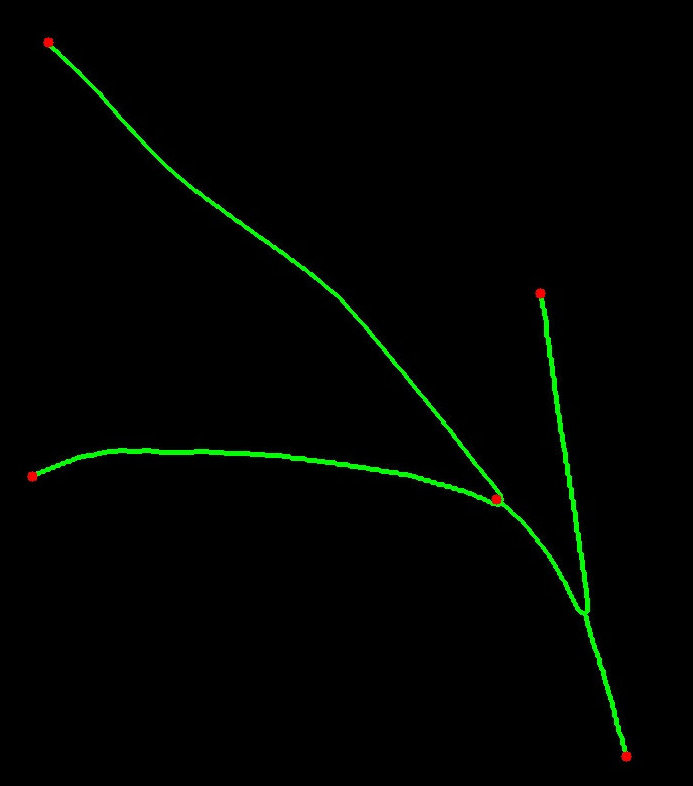}}
\caption{Extracting a skeleton for a frame}\label{fig:skel}
\end{minipage}
\begin{minipage}[t]{0.47\linewidth}\leavevmode\centering
\includegraphics[width=0.475\linewidth]{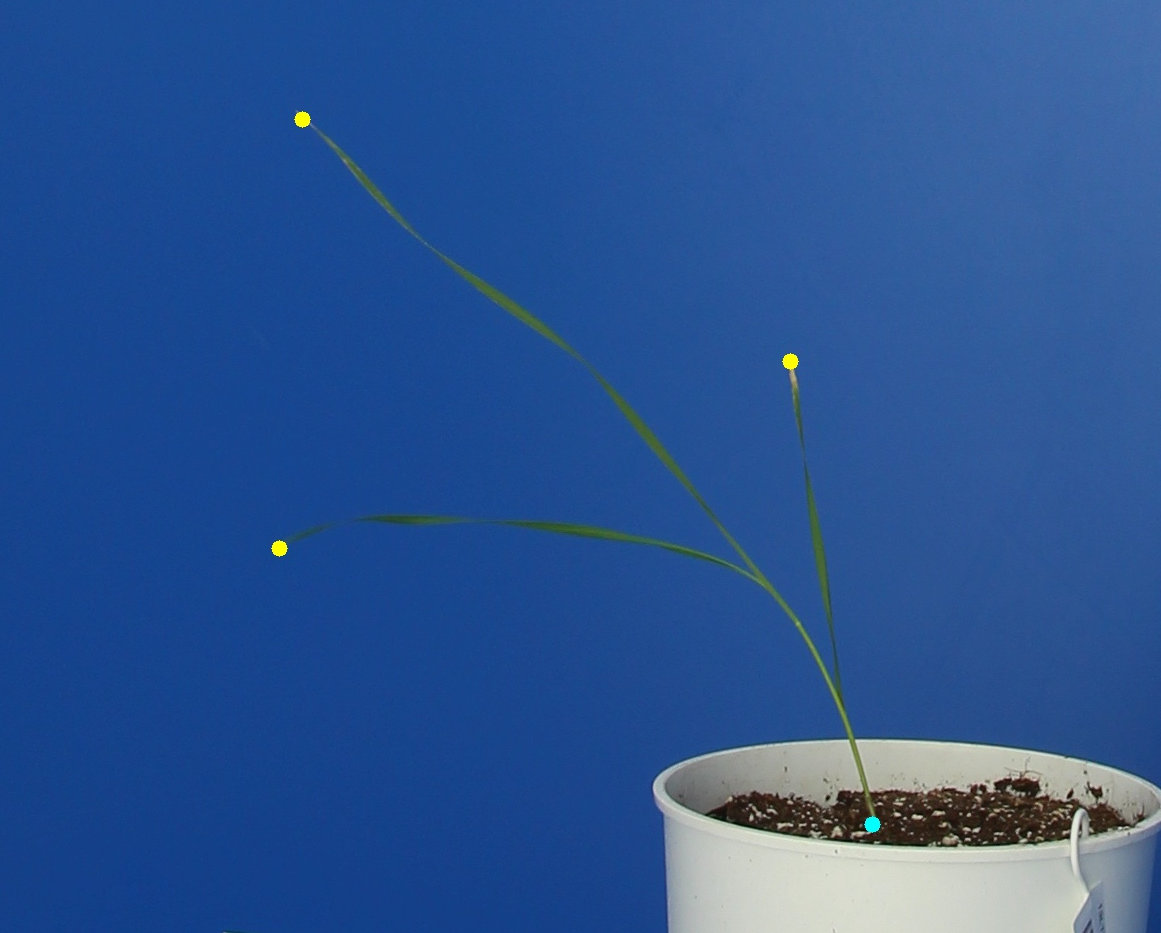}
\includegraphics[width=0.475\columnwidth]{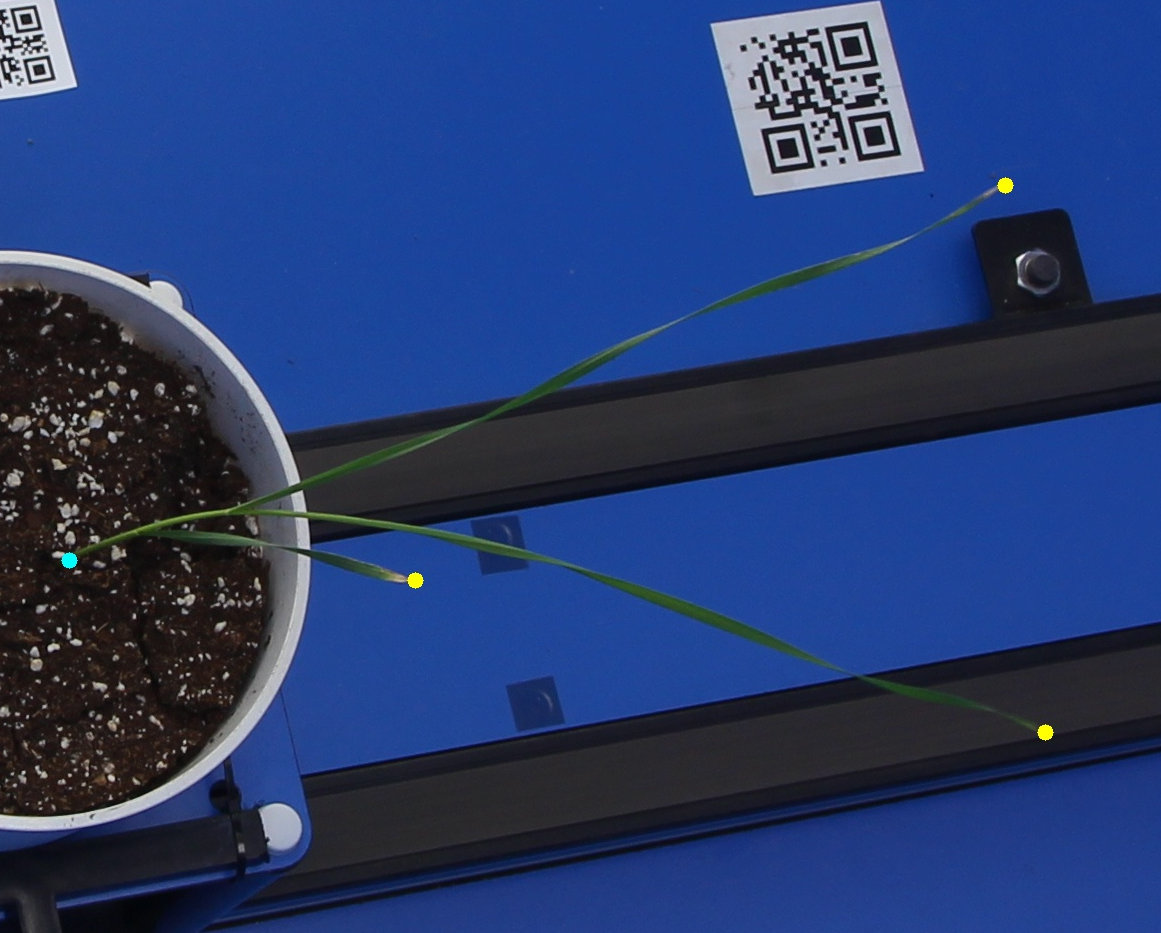}
\caption{2 views of the 3D tip and base points}
\label{fig:tips}
\end{minipage}
\end{figure}

A set of 2D skeletons extracted from the silhouette for each view are used as estimates of the projection of the set of 3D leaf axes. An example of such skeletons being used for plant reconstruction is given in the reconstruction method of \cite{OrthoImages}, where matching between skeleton points in orthographic images is used to recover 3D leaf paths. To generate the skeletons, we use the thinning algorithm of \cite{Thinning}. An example of a skeleton extracted from a silhouette is shown in Figure~\ref{fig:skel}.

\subsection{Leaf Tip Detection}

To limit the number of candidate models which need to be evaluated to find a model which corresponds to the current image set, information extracted from the 2D skeletons is used to guide the generation process. To identify possible tip points, we first construct a graph from each skeleton image. As some plant regions in the silhouettes may be disconnected due to sections of the plant which are too thin to be detected, edges are added between nearby points to connect these isolated regions in the graph.

From the graphs, we extract a set of possible 3D leaf tip points and a base point. We significantly reduce the space of possible models by considering only candidate leaf models with ends corresponding to these tip and base points. For each graph, we first detect a set of 2D points possibly corresponding to leaf tips by measuring the distance to the graph centre for each node and finding local maxima for this distance. These 2D points are matched between images to give possible 3D leaf tip locations. Matches for a point are found by locating points close to the corresponding epipolar line in a second view. 3D tip points are then determined by triangulation. Matches in further views are located by finding points close to the projection of the 3D points. The final position for each point is determined as the 3D point minimizing the sum of squared distances to the corresponding 2D points in all views. We select the 3D point closest to the input pot centre position as the base point. The selected base point and set of possible tip points for a plant are shown in Figure~\ref{fig:tips}.

The boundary of the extracted silhouette may not be smooth due to misclassified background pixels. This results in extreme points in the graph which do not correspond to leaf tips, and 3D tip points being generated corresponding to points part way along the leaves. While such points could be removed using morphological filtering operations, doing so also eliminates important structural information.

Instead, we use the set of graphs to detect 3D tip points which are likely to be part way along the path to a true leaf tip. We find the shortest path in the graph from each 3D tip point to the base point in each view where that point is visible, and remove any point for which these paths do not include at least $150$ pixels not included in the paths to a tip point farther from the base. Points with 2D projections which are not close to the silhouettes are also rejected.

\subsection{Leaf Generation}\label{sec:leafgen}

For each possible tip point, we build a set of candidate leaf models. Leaves from the database of manually modelled plants are linearly transformed to fit the tip and base points for each leaf to the selected tip point and base point positions. The transformed leaves are then evaluated against the images. The leaf models which best match the images are determined by measuring the distance in each image between sampled points on the models and the nearest point on the 2D skeleton for that image.

The tip point, base point, and the orientation vector determined in the calibration process are used to define a linear transform mapping the base and tip of each leaf chosen from the database to the corresponding points in the current scene. This transform can then be used to map all points of the leaf into the scene. To efficiently evaluate distances from model points to skeleton points, a distance transform is applied to the 2D skeleton in each view, assigning each pixel in the image the distance to the nearest skeleton point. As leaves can hang over the edge of the pot, where they cannot be seen by some cameras, we also make use of a 3D model giving the approximate structure of the pot and pot holder. This allows occlusion to be estimated and incorporated into the reconstruction. After evaluating the full set of transformed leaf models against the images, parameters for the best models are refined to improve their fit to the image set, as we do not expect the database to contain an exact match for each leaf.

\begin{figure}[t]
\centering
\subfigure[with penalty]{\includegraphics[width=0.25\textwidth]{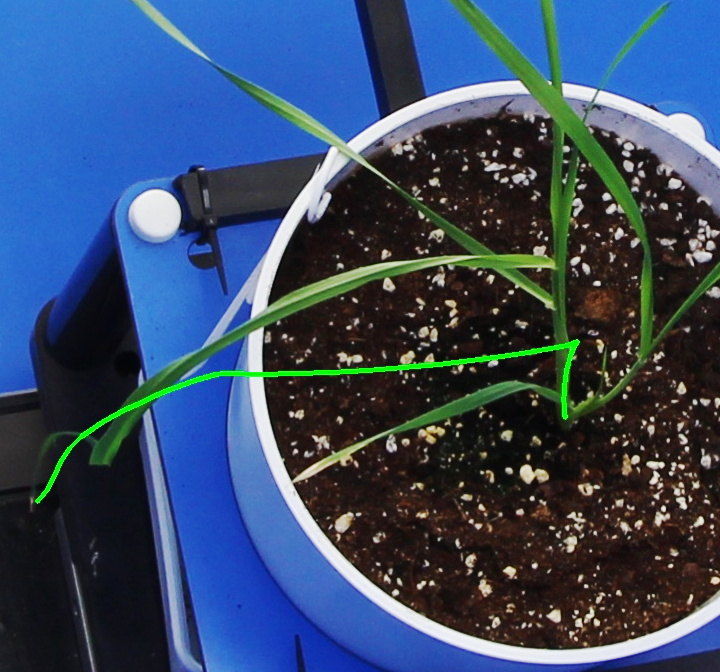}}
\subfigure[without penalty]{\includegraphics[width=0.25\textwidth]{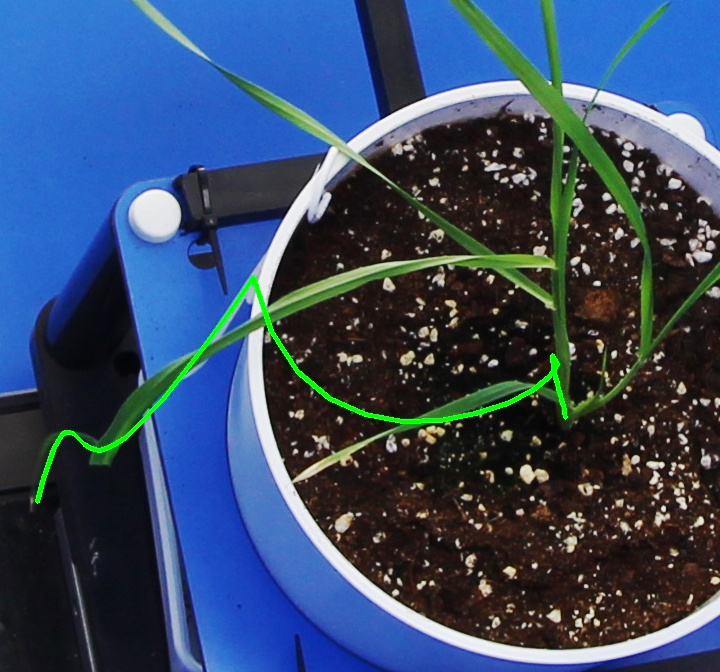}}
\caption{The effect of the curvature penalty}\label{fig:curvepenalty}
\end{figure}

To refine the leaf while preserving its shape, we model each leaf
using cubic B-splines $b_\mat{C}(t)\rightarrow\Re^3$, $t\in(0,1)$
parameterised by a set of control points $\vec{C}$.  The control
points are optimised with respect to ${\cal S}=\{ \mat{S}^v\}_{\forall
  v}$, where $\mat{S}^v=[ \vec{s}_1^v, \hdots, \vec{s}_n^v]$ is the
set of skeleton points in view $v$, by minimising
\begin{equation}
d(\mathcal{S}, \mat{C})=\sum_v
\int_0^1 \frac{r_v(b_{\mat{C}}(t))
}{
\sum_v o_v(b_{\mat{C}}(t))
}\;\mathit{dt}+\int_0^1 c_{\mat{C}}(t) dt
\label{eqn:leafoptlm}\end{equation}
where the residual
\begin{equation}
r_v(\vec{x})=o_v(\vec{x})\left(\min_j\|\vec{s}_j^v-\mat{A}_v\vec{x}\|_2\right)\label{eqn:distresid}
\end{equation}
measures the distance between the projection of a point on the leaf
against the closest skeleton point in view $v$.  Here, $\mat{A}_v$ is
the projection matrix for view $v$ and $o_v(\vec{x})$ is a delta
function that is $0$ if $\vec{x}$ is occluded in view $v$ and $1$ otherwise. Residuals are inversely weighted by the number of views where a point is visible, to avoid biasing the optimisation towards a better fit for points which are visible in more views. To prevent significant changes in the leaf shape, the term
\begin{equation}
c_{\mat{C}}(t)=\alpha(\kappa_{\mat{C}}(t)-\kappa_{\mat{C}_0}(t))^2
\end{equation}
is added to the residuals to penalise changes in curvature with respect
to the control points $\mat{C}_0$ of the original curve. The term
\begin{equation}
\kappa_{\mat{C}}(t)=
\frac{\| b_{\mat{C}}'(t)\times b_{\mat{C}}''(t) \|}
     {\| b_{\mat{C}}'(t) \|^{3}}
\end{equation}
measures curvature of the B-spline $b_{\mat{C}}$ evaluated at $t$.
The effect of the curve penalty on the reconstruction is illustrated
in Figure \ref{fig:curvepenalty}.  In both cases the optimisation
began from the curve illustrated in Figure \ref{fig:curvepenalty}(a).
Without the penalty, the different parts of the hypothesised curve
latch onto different, disjoint leaves in the image (Figure
\ref{fig:curvepenalty}(b)).

The curvature of a leaf may not be continuous, particularly where the
leaf meets the stem. We therefore find any points in the 3D path of the original leaf model where a sharp change of angle ($>45^\circ$) occurs, and model the path as a set of one or more
connected splines, with discontinuous curvature at these points. The
number of control points for each segment is determined from the
segment length.

\begin{figure}[t]
\leavevmode\centering
\includegraphics[width=0.2435\textwidth]{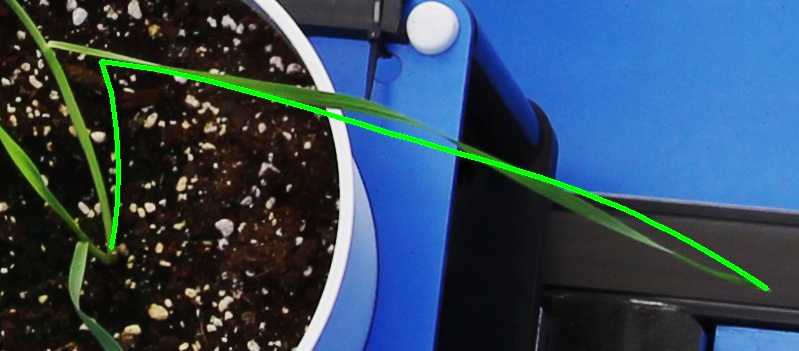}
\includegraphics[width=0.2435\textwidth]{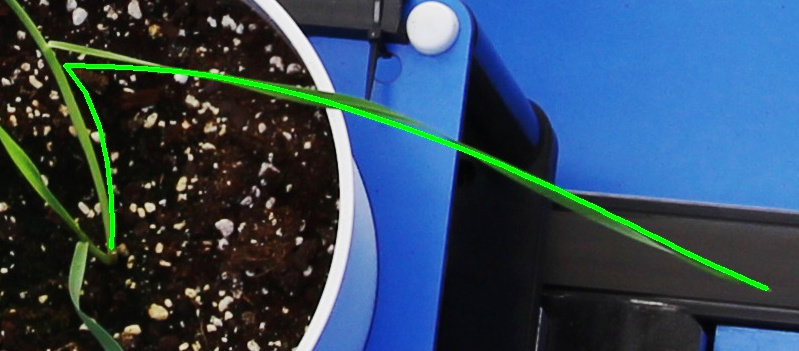}
\includegraphics[width=0.2435\textwidth]{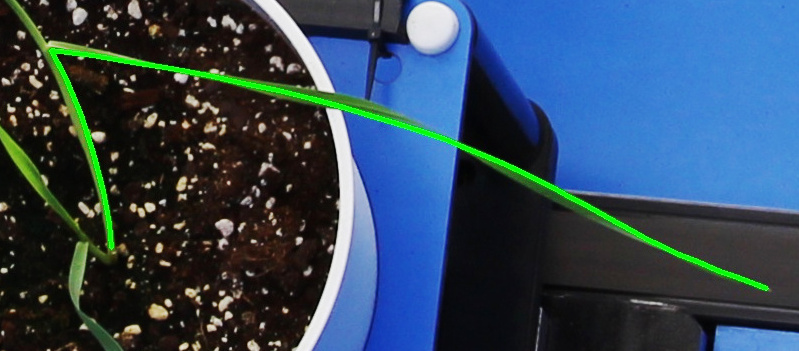}
\caption{Refining parameters for a leaf model}
\label{fig:leaf2}
\end{figure}

To fit a leaf to the image set, the position of the 3D control points
are refined by applying Levenberg-Marquardt optimisation to a set of
points sampled along the leaf.  In practice, we define $n$ points that
are separated by approximately $7.5\mathrm{mm}$ along the original
curve and define the distance residual in \ref{eqn:distresid} by the
distance transform over the skeletonised observation.  The change in
the shape of a leaf during refinement is illustrated in
Figure~\ref{fig:leaf2}.

\begin{figure}[t]
\leavevmode\centering
\subfigure[Initial set]{\includegraphics[width=0.37\textwidth]{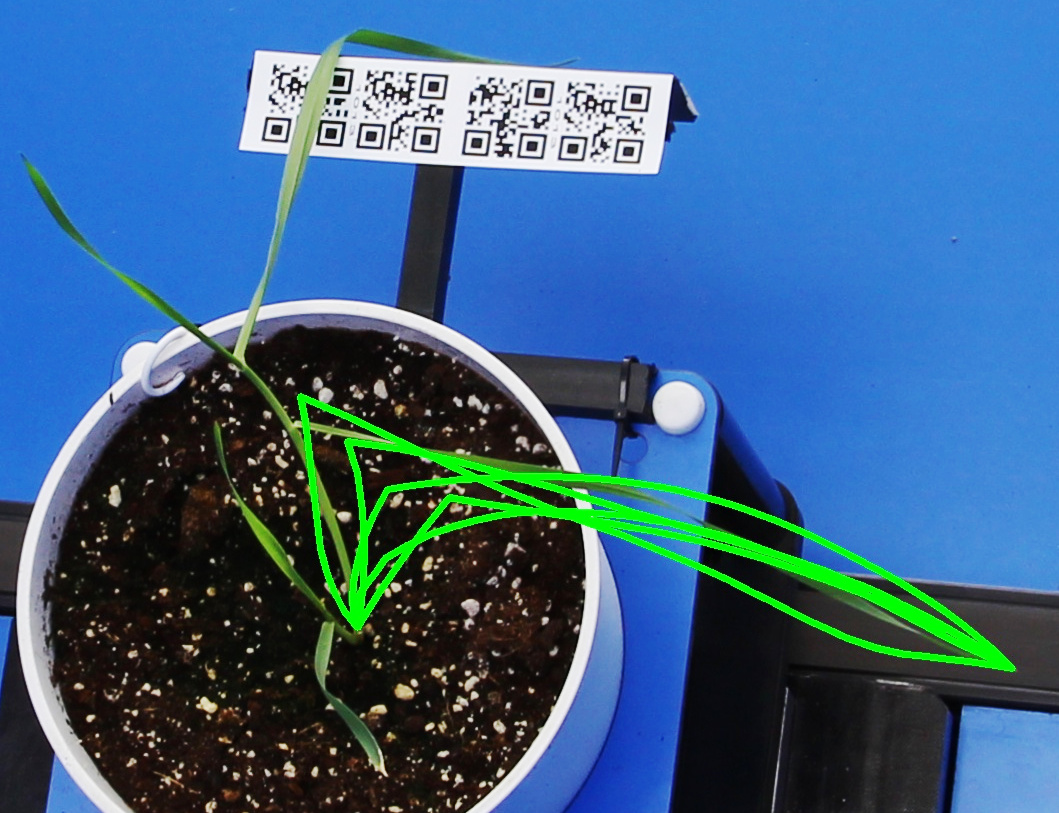}}
\subfigure[Optimised set]{\includegraphics[width=0.37\textwidth]{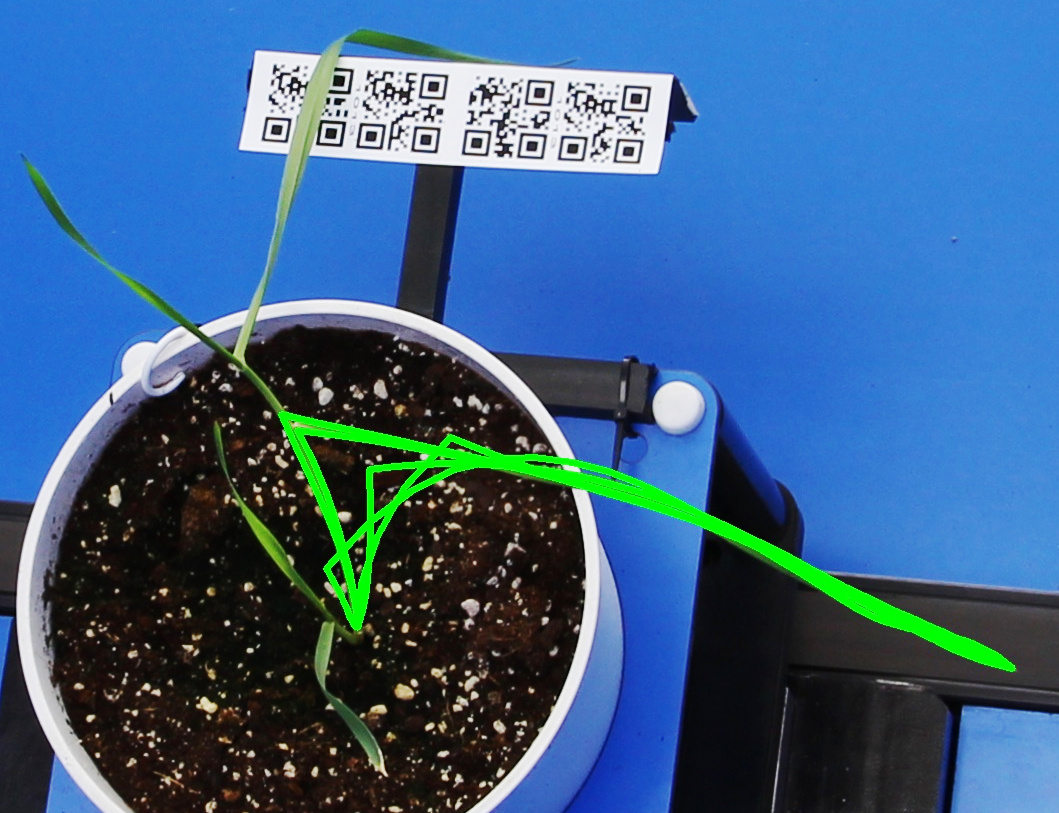}}
\caption{Initial and optimised set of leaf candidates}
\label{fig:leaf3}
\end{figure}

The distance measure is used to rank the full set of leaf models generated from the database. The best $200$ leaf models are then selected and refined. Figure~\ref{fig:leaf3} shows a set of initial candidate models obtained for a point, and the same set of models after refinement.

\subsection{Structure Estimation}

\begin{figure}[t]
\centering
\includegraphics[width=0.25\textwidth]{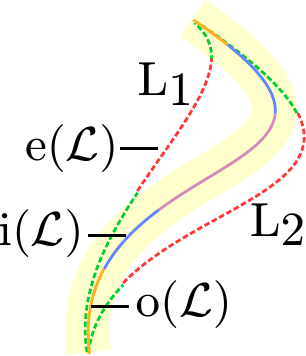}
\vspace*{1.25mm}
\caption{The interior, exterior and overlapping areas of the set metric}
\label{fig:leafsetmetric}
\end{figure}

The above  process generates a set of possible leaf models which may be combined into a full plant model. In generating the complete plant model, multiple candidates for each tip are tested, because overlapping leaves may result in several plausible paths from a tip to the base. For each tip point, we select $5$ candidate leaves using the distance measure evaluated for the refined leaf. As multiple leaf models may converge to the same shape in refinement, additional leaf models are not selected if there is only minimal deviation from an already selected model.

On the basis of the leaf hypothesis set, and the anatomy-based prior which describes the ways in which such leaves may be combined, it is possible to construct a set of full-plant hypotheses. This process may be seen as a data-driven means of exploiting a generative model in a situation where sampling from a full generative model directly would be too computationally expensive.  The generative model for a plant such as wheat is relatively simple, but nonetheless far too complex to be sampled from directly.

\newcommand{\LeafRender}{{\cal R}}

Each hypothesised structure is evaluated against the number of
skeleton pixels covered by the model, the number of pixels outside the
plant which are covered, and the number of leaves used.  Let
$\Silhouette_v$ be the set of skeleton pixels in view $v$. The set of
\emph{`interior'} pixels which are supported by the set of leaves
$\cal L$ is given by
\begin{equation}
i_v({\cal L})=\{ \vec{i} \;|\; (\vec{i}\in\Silhouette_v) \wedge (a_v(\vec{i}, {\cal L})>0) \}
\end{equation}
where
\begin{equation}
a_v(\vec{i}, {\cal L})=\sum_{\mat{L}\in{\cal L}} m_v(\vec{i}, \mat{L})
\end{equation}
and
\begin{equation}
m_v(\vec{i}, {\mat L})=\left\{\begin{array}{cl}
1 & \textrm{if\;} \min_{t}  \| \vec{i}-\mat{A}_vb_{\mat{L}}(t) \| < \tau \\
0 & \textrm{otherwise.}
\end{array}\right.
\end{equation}
counts the number of leaves which project to the pixel
$\vec{i}\in\Silhouette_v$ within a tolerance $\tau=10$ pixels.  This
threshold helps to account for divergence between the skeleton
extracted for each frame and the projection of the true axis of each
leaf. The set of \emph{`exterior'} pixels
\begin{equation}
e_v({\cal L})=\{ \vec{j} \;|\; (\vec{j}\in\LeafRender) \wedge (\min_{\vec{i}\in\Silhouette_v} \|\vec{i}-\vec{j}\| > \tau) \}
\end{equation}
are in the image $\LeafRender$ generated by rendering ${\cal L}$ with
projection matrix $\mat{A}_v$ but are not within the threshold distance of any skeleton pixels.
The quality of the leaf set is
\begin{equation}
q({\cal L})=\sum_{v\in{\cal V}} \left(|i_v({\cal L})|-\beta|e_v({\cal L})|-\gamma o_v({\cal L})\right)\label{eqn:leafsetmetric}
\end{equation}
where $\beta$ controls the penalty for covering exterior pixels and $\gamma$ with
\begin{equation}
o_v({\cal L})=\sum_{\vec{i}\in i_v({\cal L})} (a_v(\vec{i}, {\cal L})-1)
\end{equation}
penalises solutions where multiple leaves overlap the same set of
pixels.  Consequently, \eqref{eqn:leafsetmetric} favours models that
closely match the skeletons in each view while using the smallest
number of leaves.  Figure \ref{fig:leafsetmetric} illustrates the
segmentation of the observed image into interior, exterior and
overlapping pixels given a hypothesised leaf set ${\cal L}$.

A set of leaves ${\cal L}$ is chosen from a larger set of candidate
leaves by a random, greedy search.  Let ${\cal
  C}^l=\{\mat{C}_1^l, \hdots, \mat{C}_n^l\}$ be the set of $n$
candidate control point sets for leaf tip $l$.  A leaf model $\mat{C}'$ is
randomly chosen and removed from the set ${\cal P}=\{ {\cal
  C}^l\}_{\forall l}$.  If $q({\cal L}\cup\{\mat{C}'\})>q({\cal L})$,
then $\mat{C}'$ is added to the initially empty set of hypothesised
leaves ${\cal L}$ and ${\cal C}^l\rightarrow\varnothing$ where
${\cal C}^l$ is the set of candidate leaves that contained
$\mat{C}'$.  The process of sampling leaves from ${\cal P}$ and adding
them to the model set continues until ${\cal P}$ is empty.

\section{Results}

\begin{figure}[!tb]
\centerline{
\includegraphics[width=0.17\columnwidth]{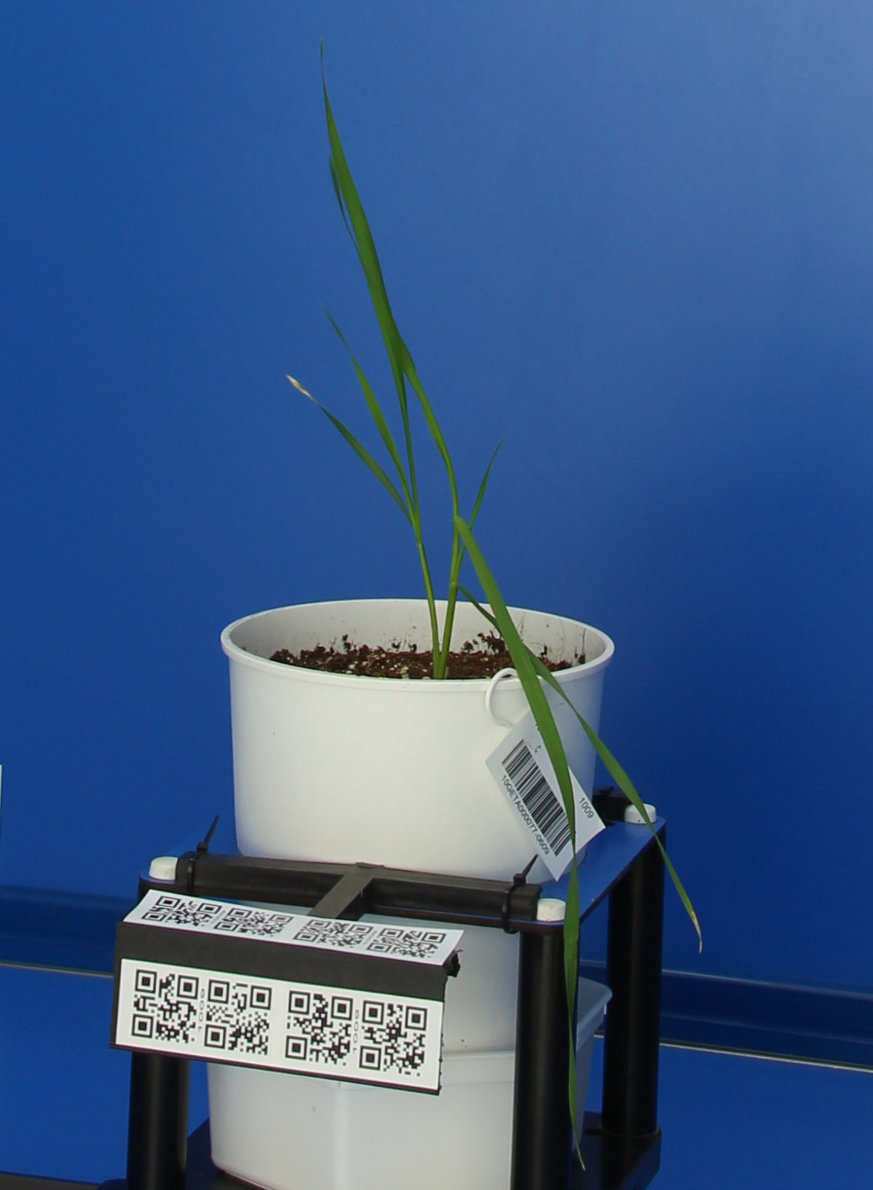}
\includegraphics[width=0.17\columnwidth]{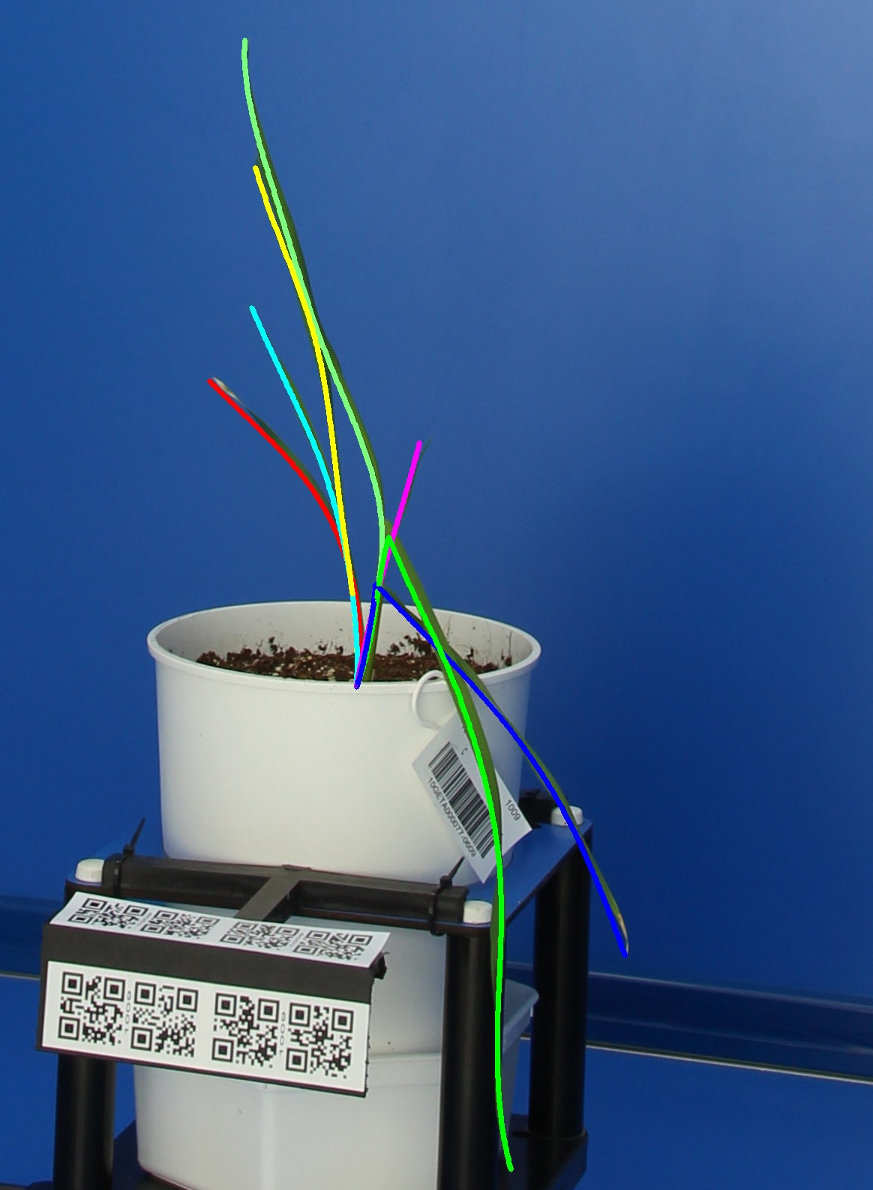}
\includegraphics[width=0.17\columnwidth]{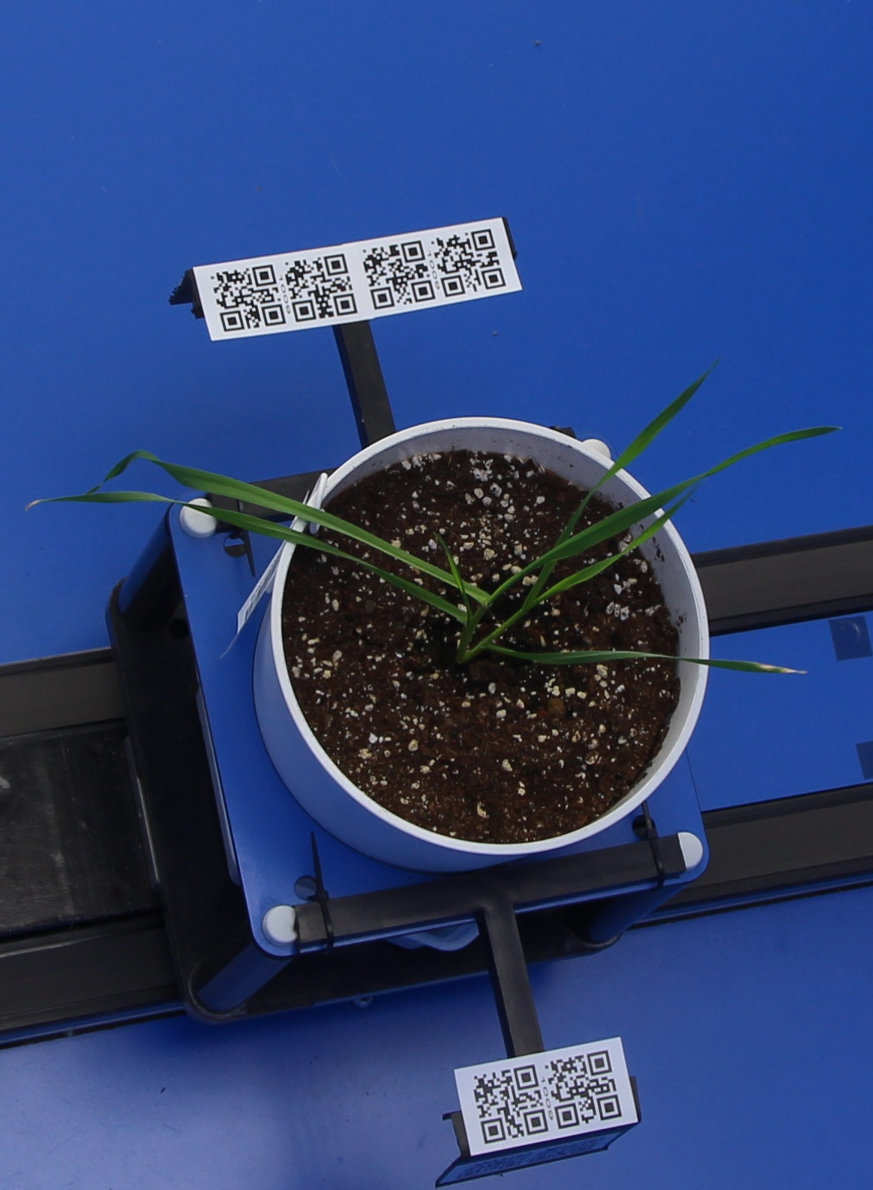}
\includegraphics[width=0.17\columnwidth]{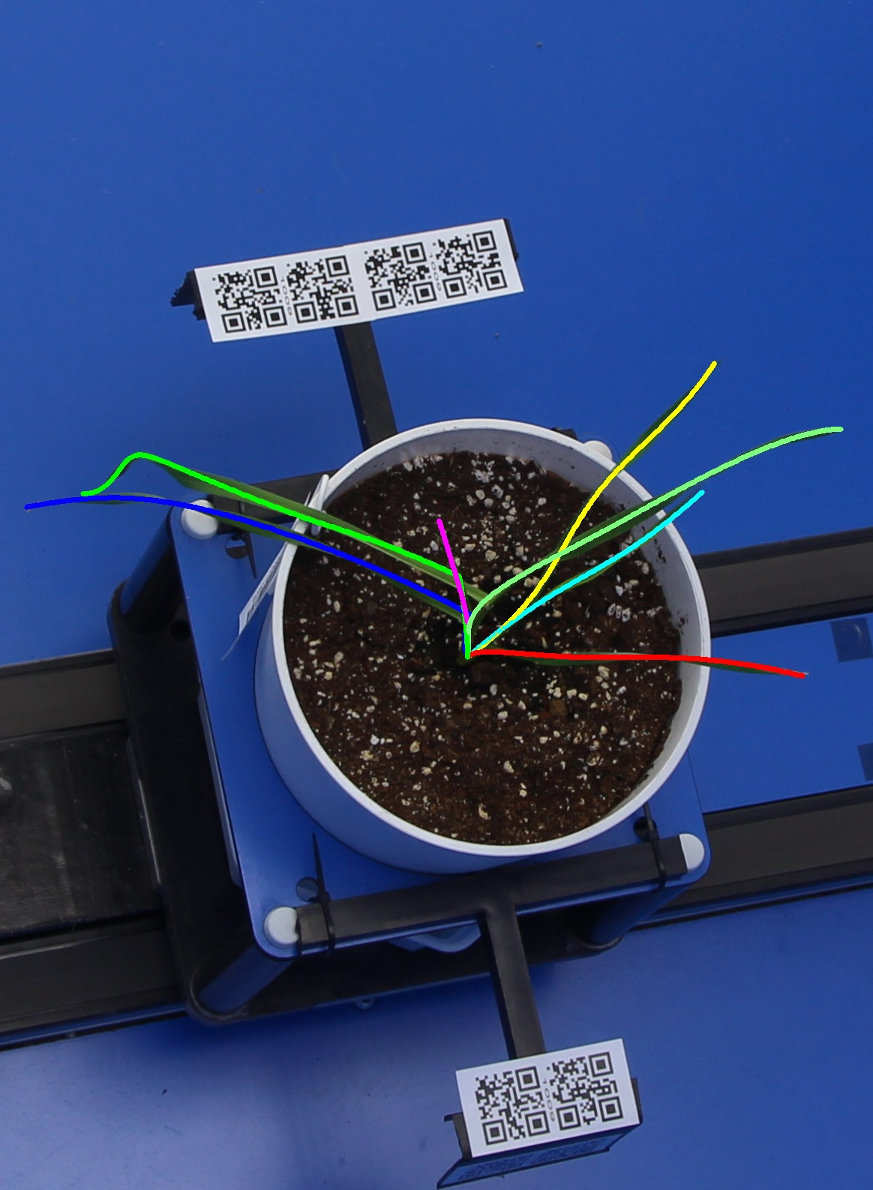}
}
\vspace{1mm}
\centerline{
\includegraphics[width=0.17\columnwidth]{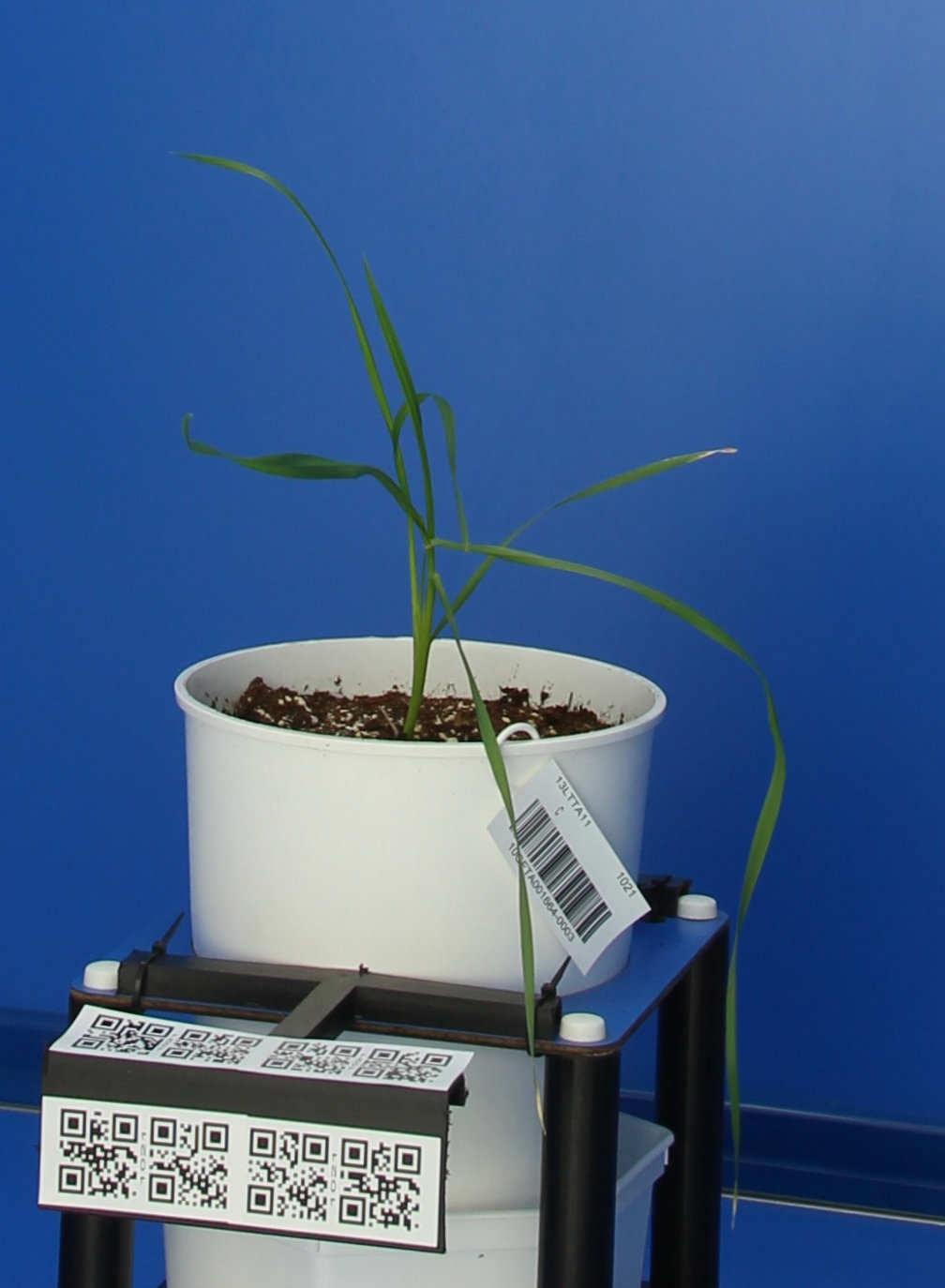}
\includegraphics[width=0.17\columnwidth]{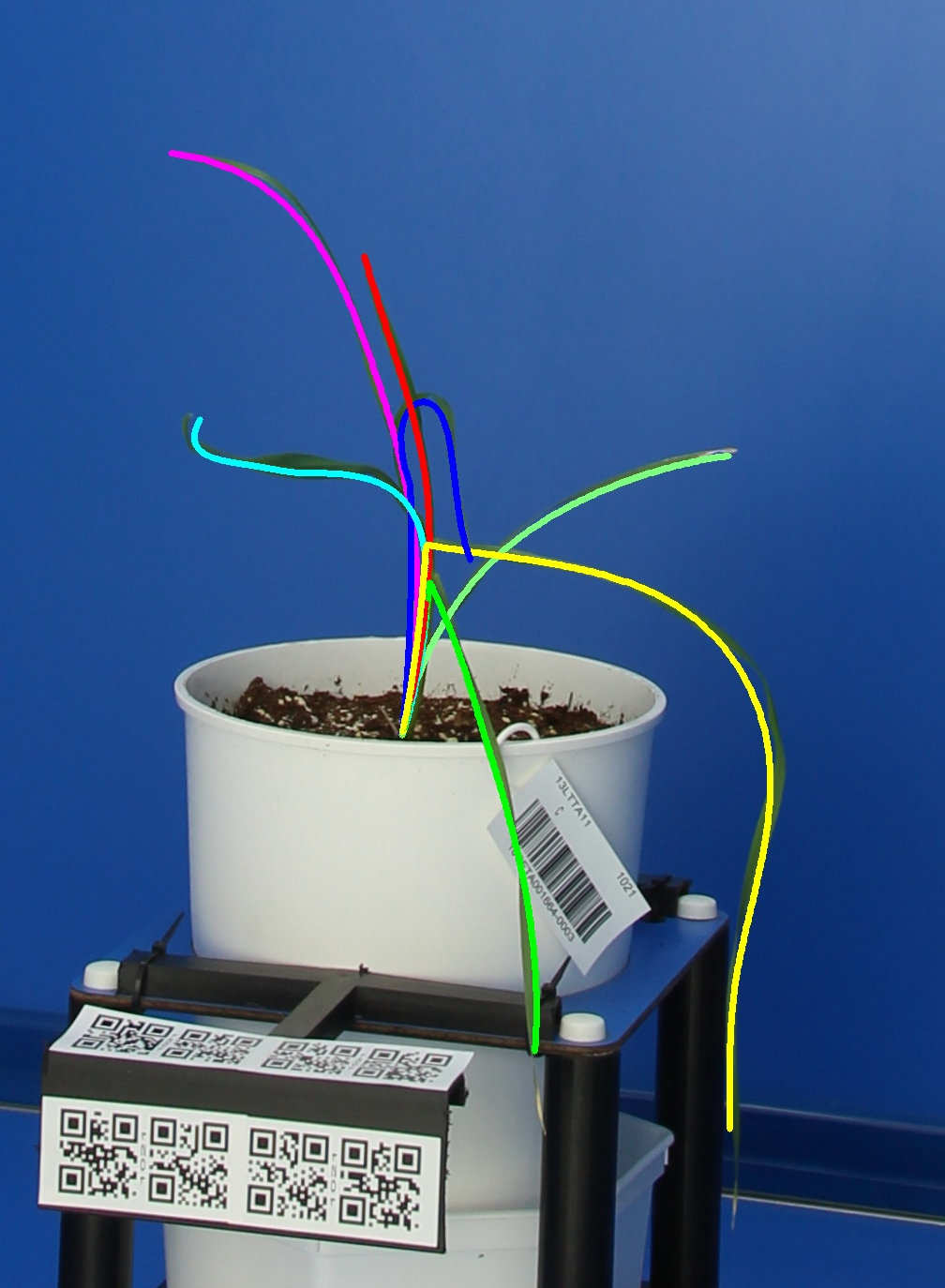}
\includegraphics[width=0.17\columnwidth]{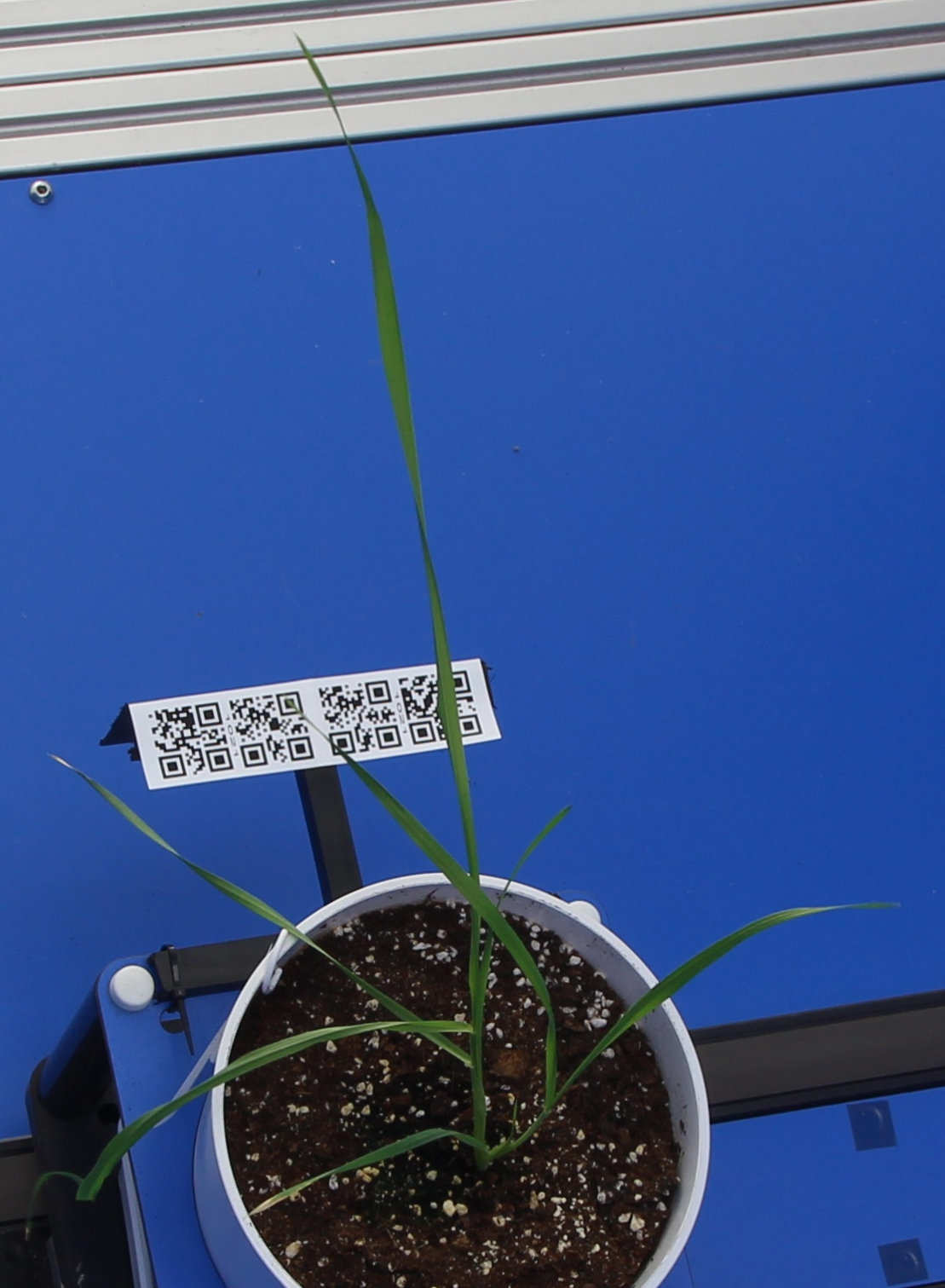}
\includegraphics[width=0.17\columnwidth]{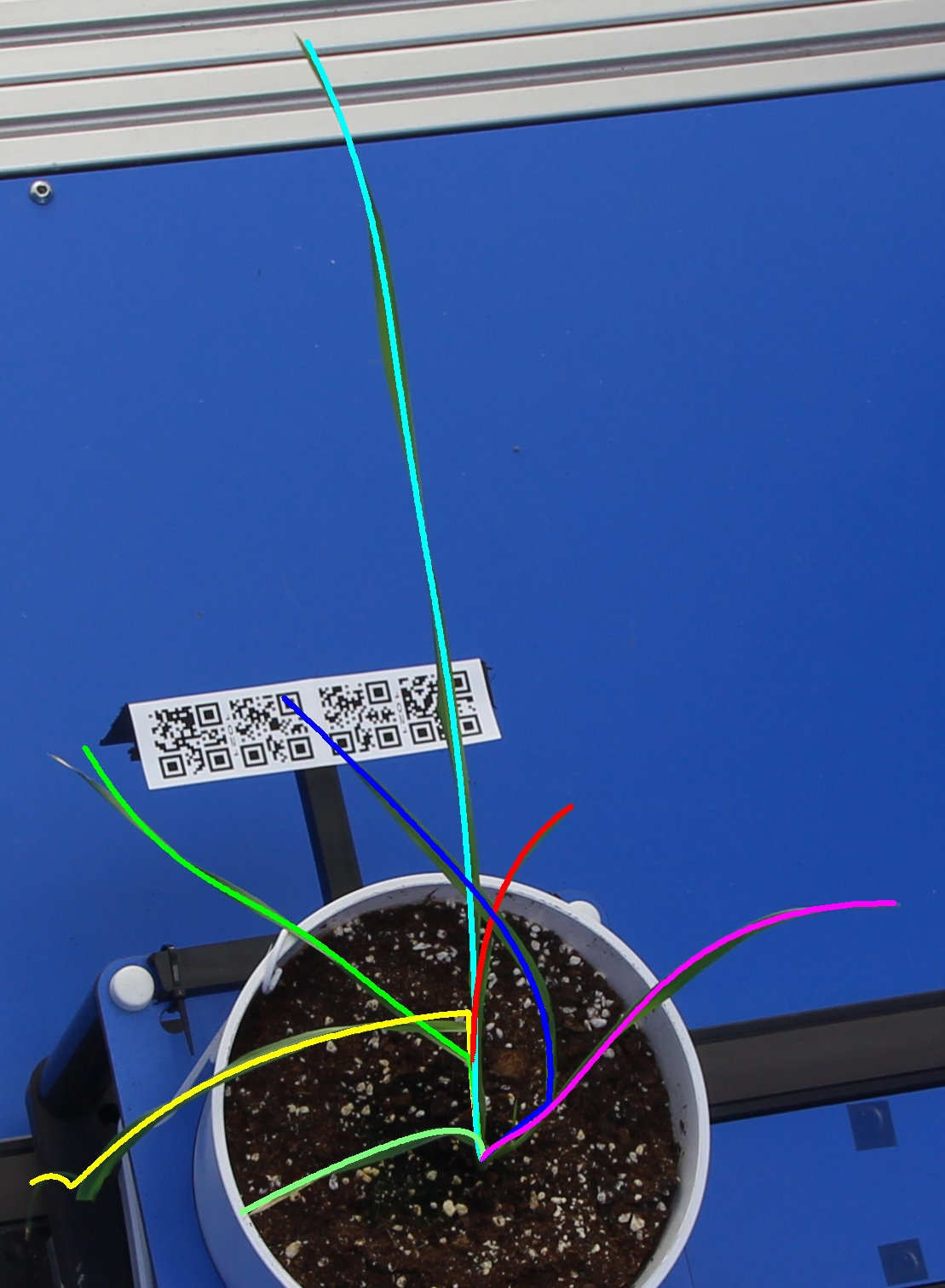}
}
\vspace{1mm}
\centerline{
\includegraphics[width=0.17\columnwidth]{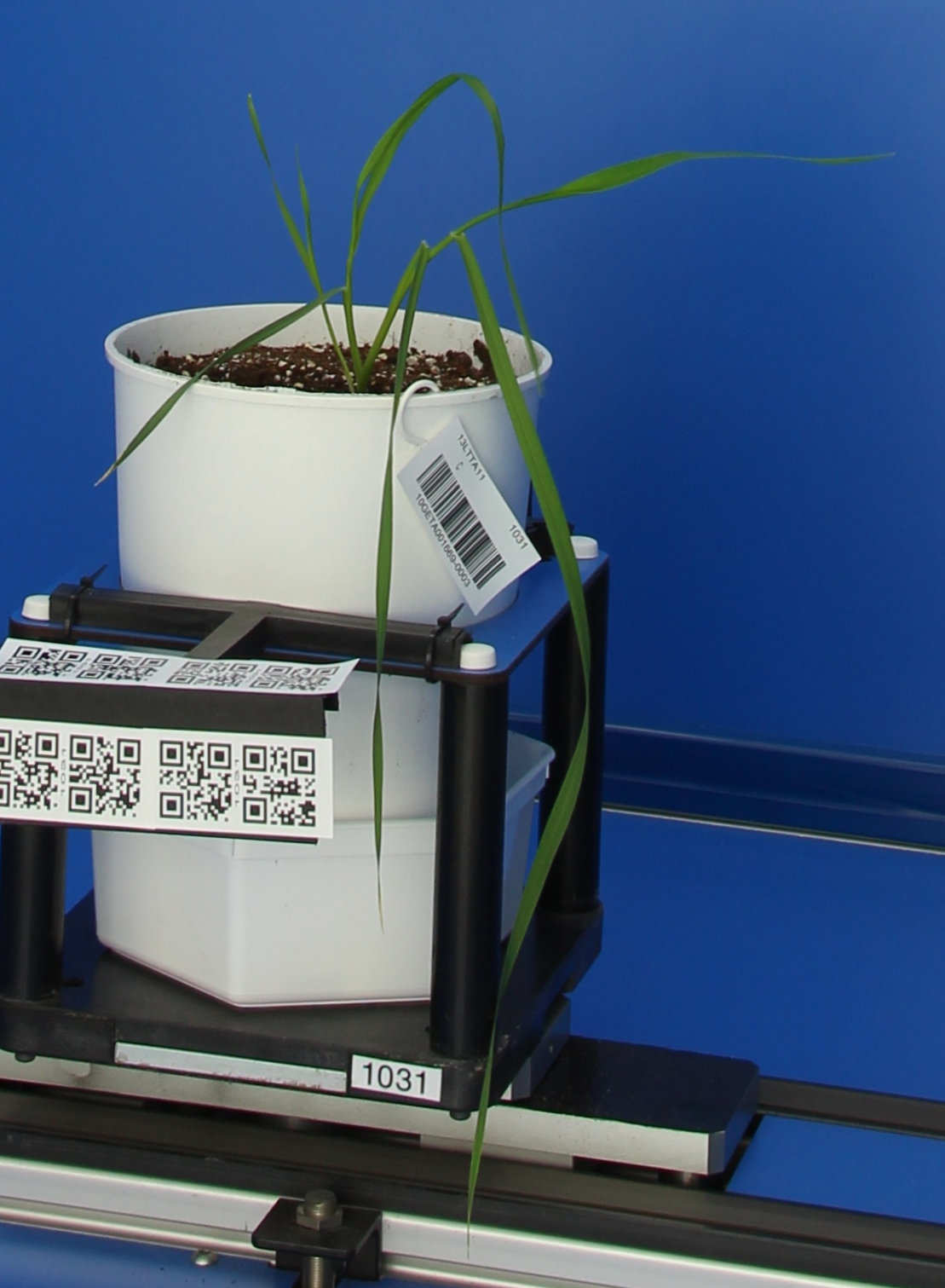}
\includegraphics[width=0.17\columnwidth]{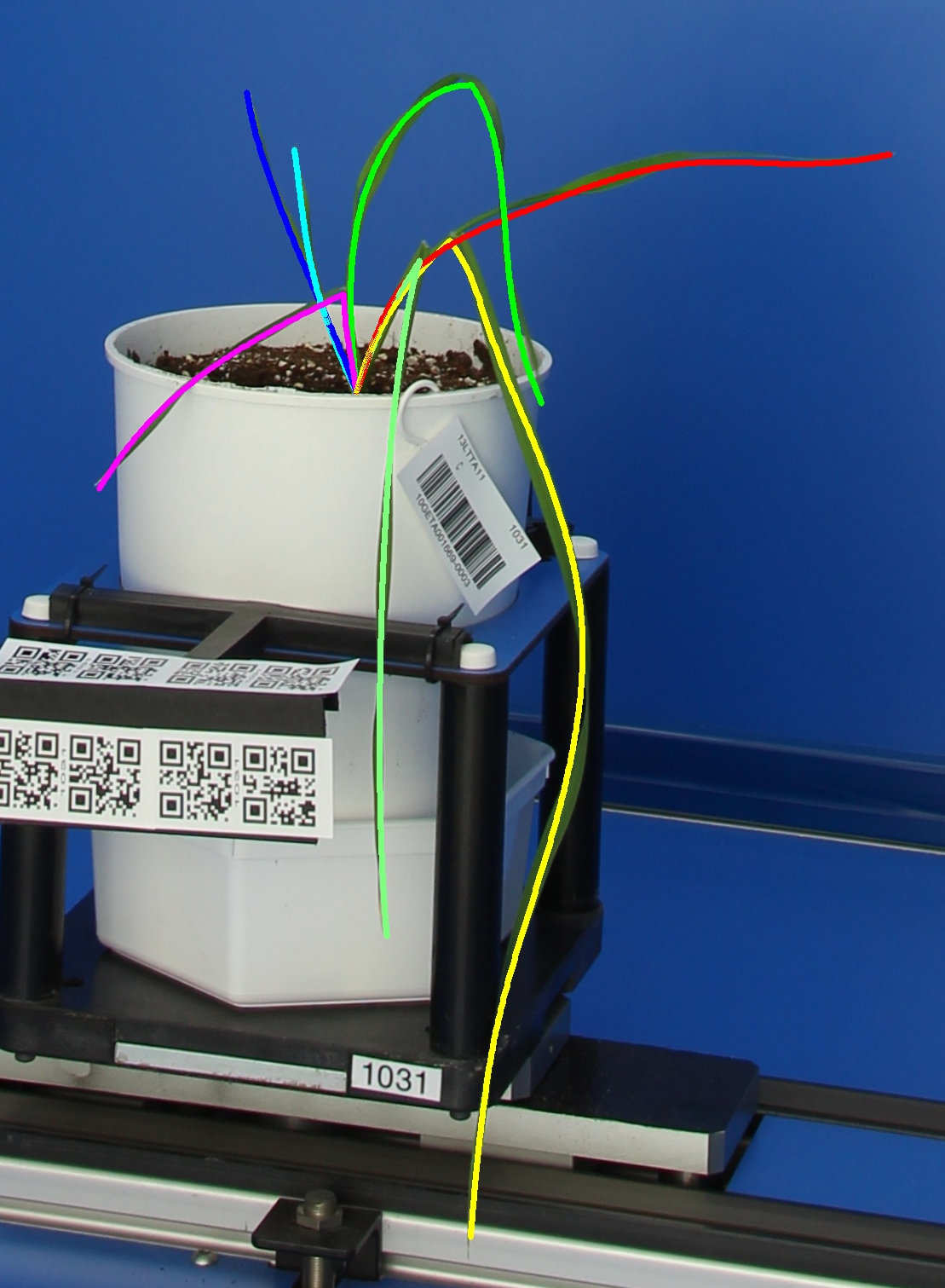}
\includegraphics[width=0.17\columnwidth]{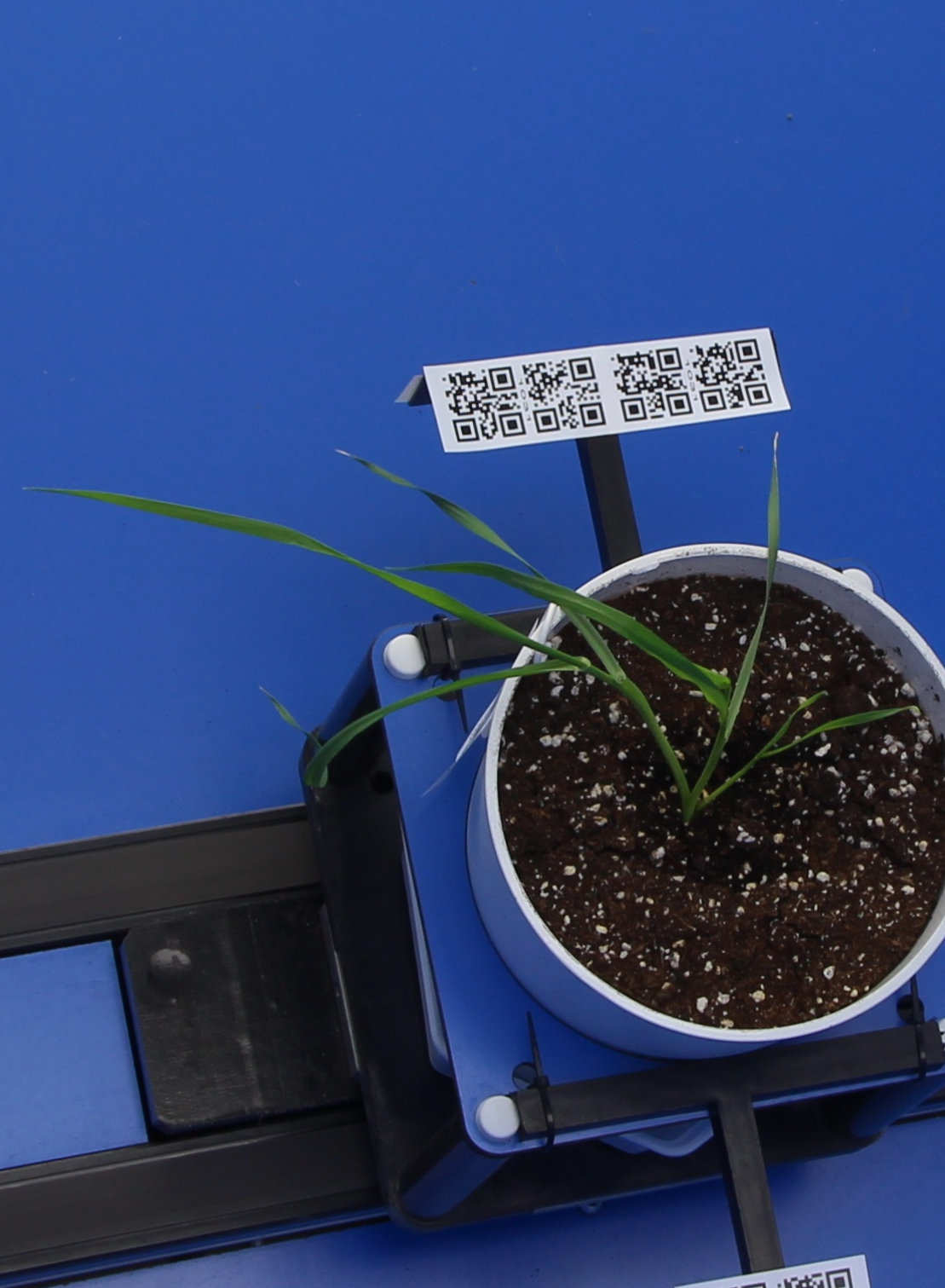}
\includegraphics[width=0.17\columnwidth]{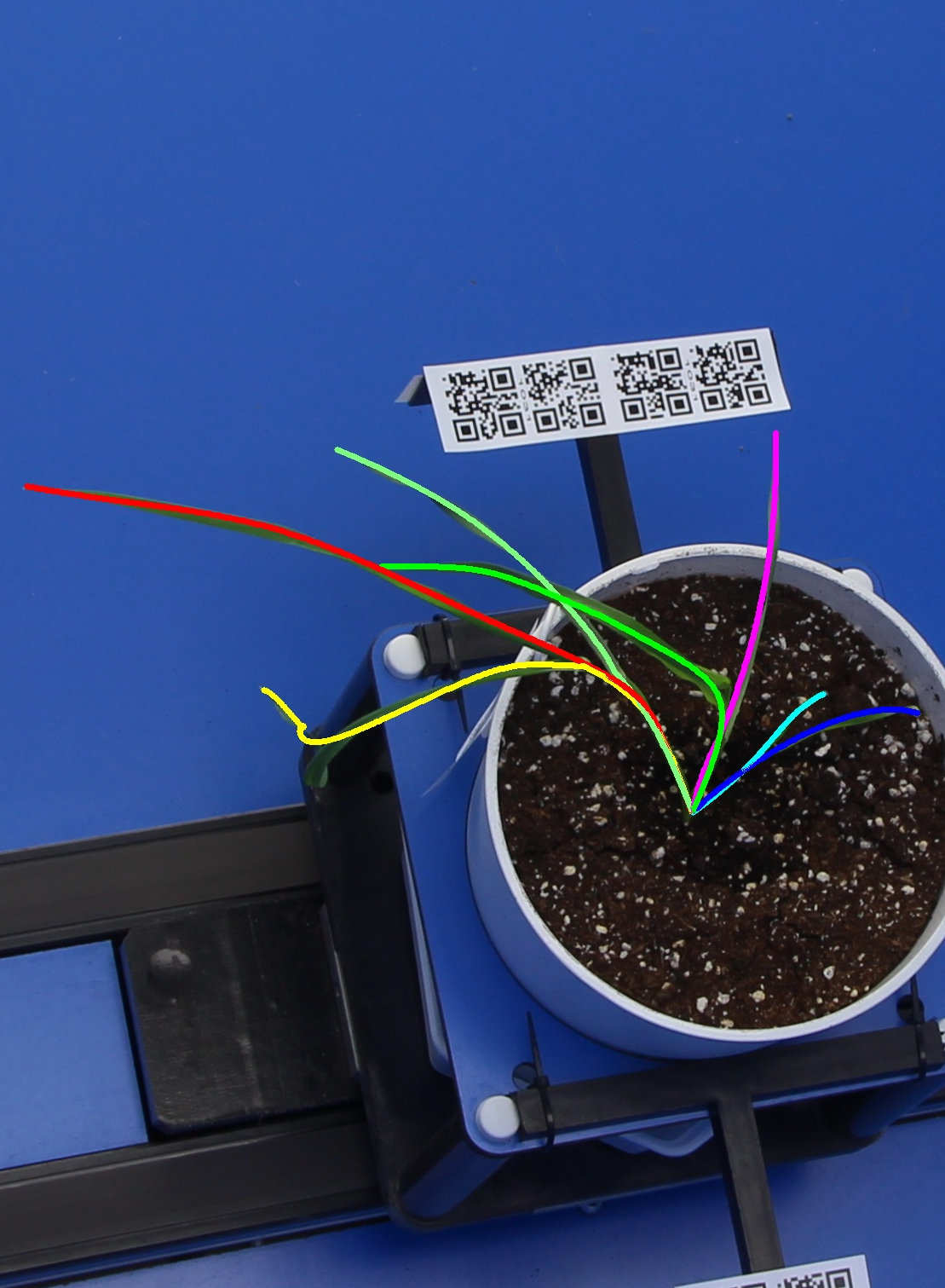}
}
\vspace{1mm}
\centerline{
\includegraphics[width=0.17\columnwidth]{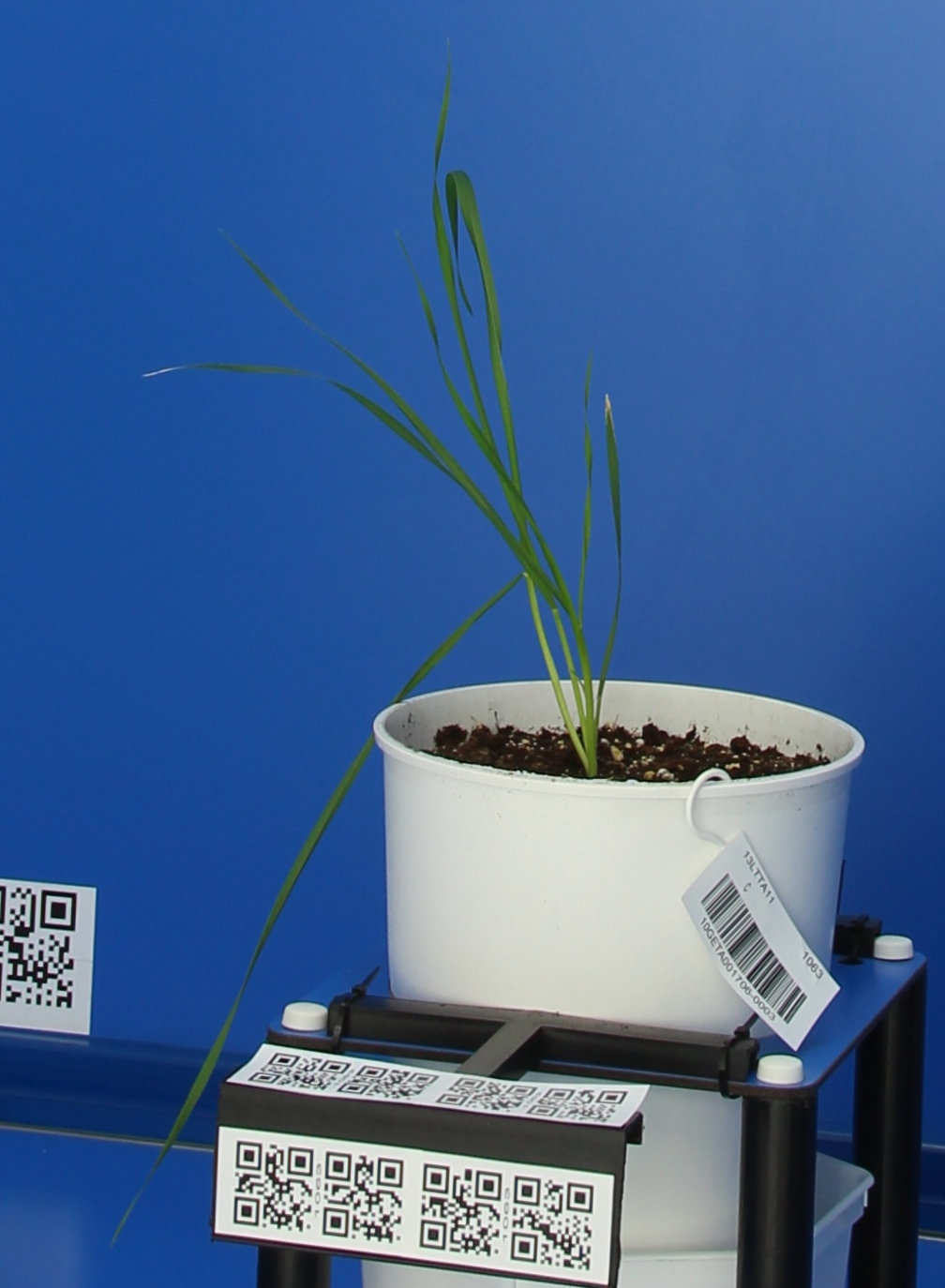}
\includegraphics[width=0.17\columnwidth]{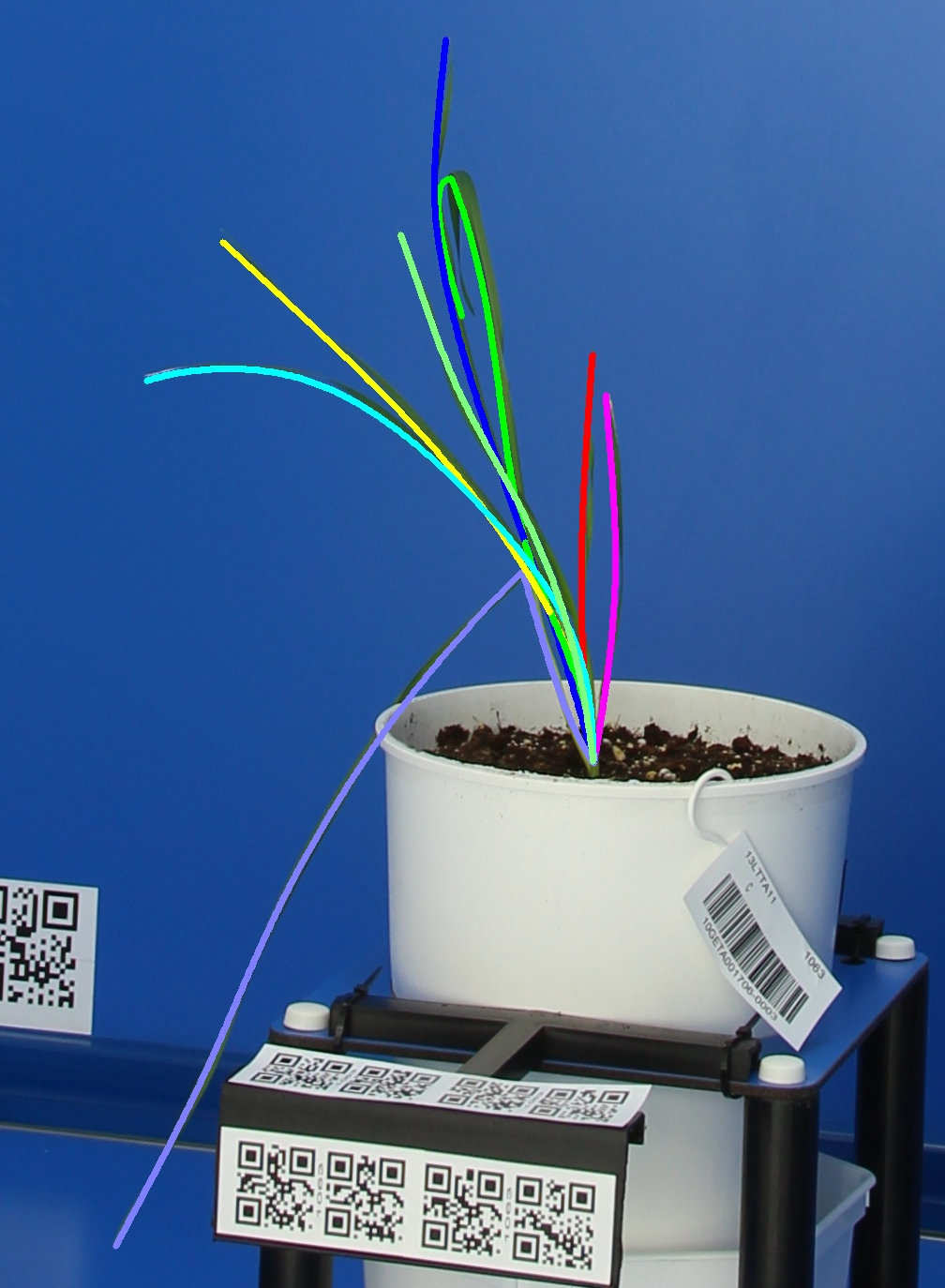}
\includegraphics[width=0.17\columnwidth]{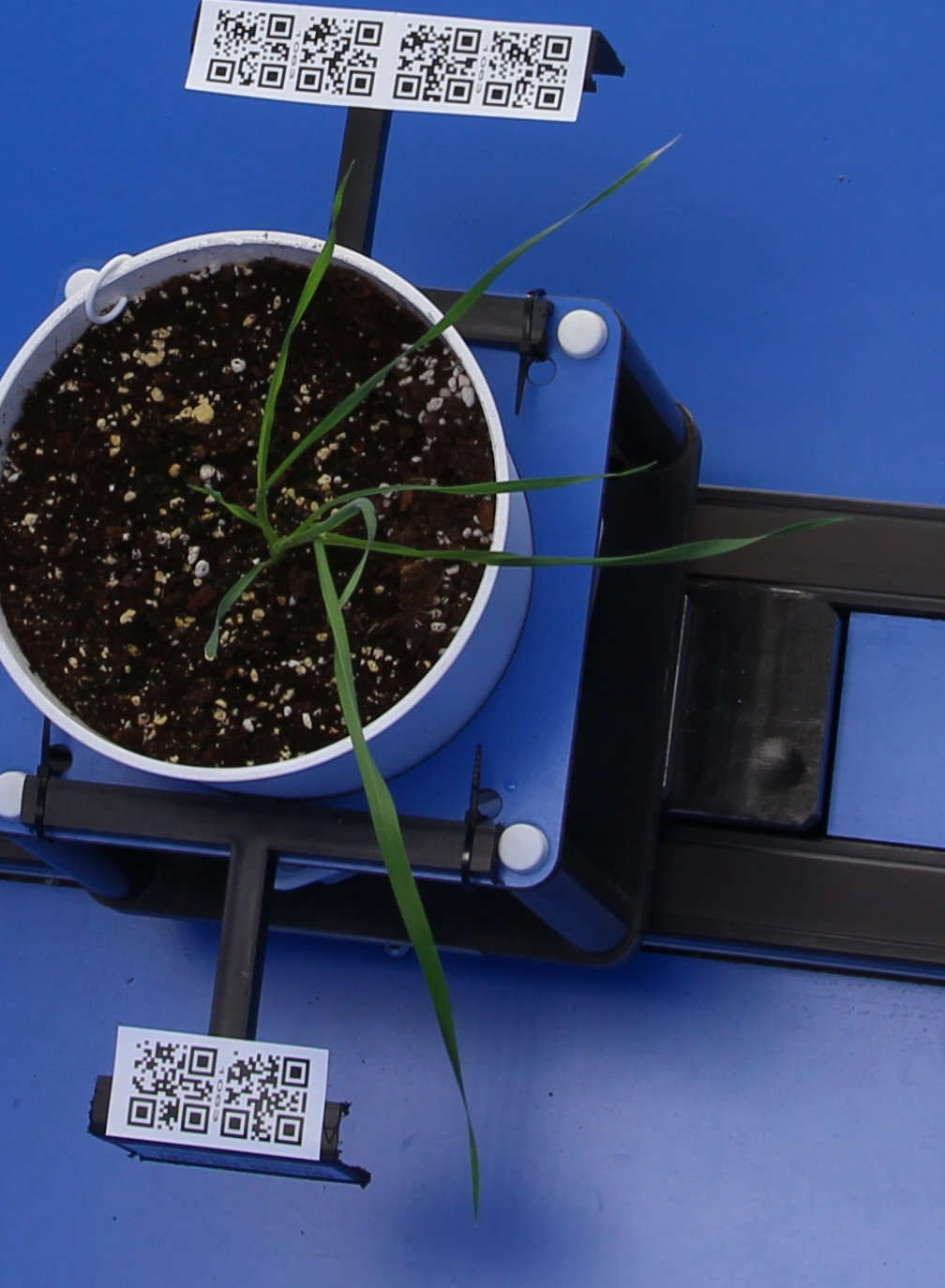}
\includegraphics[width=0.17\columnwidth]{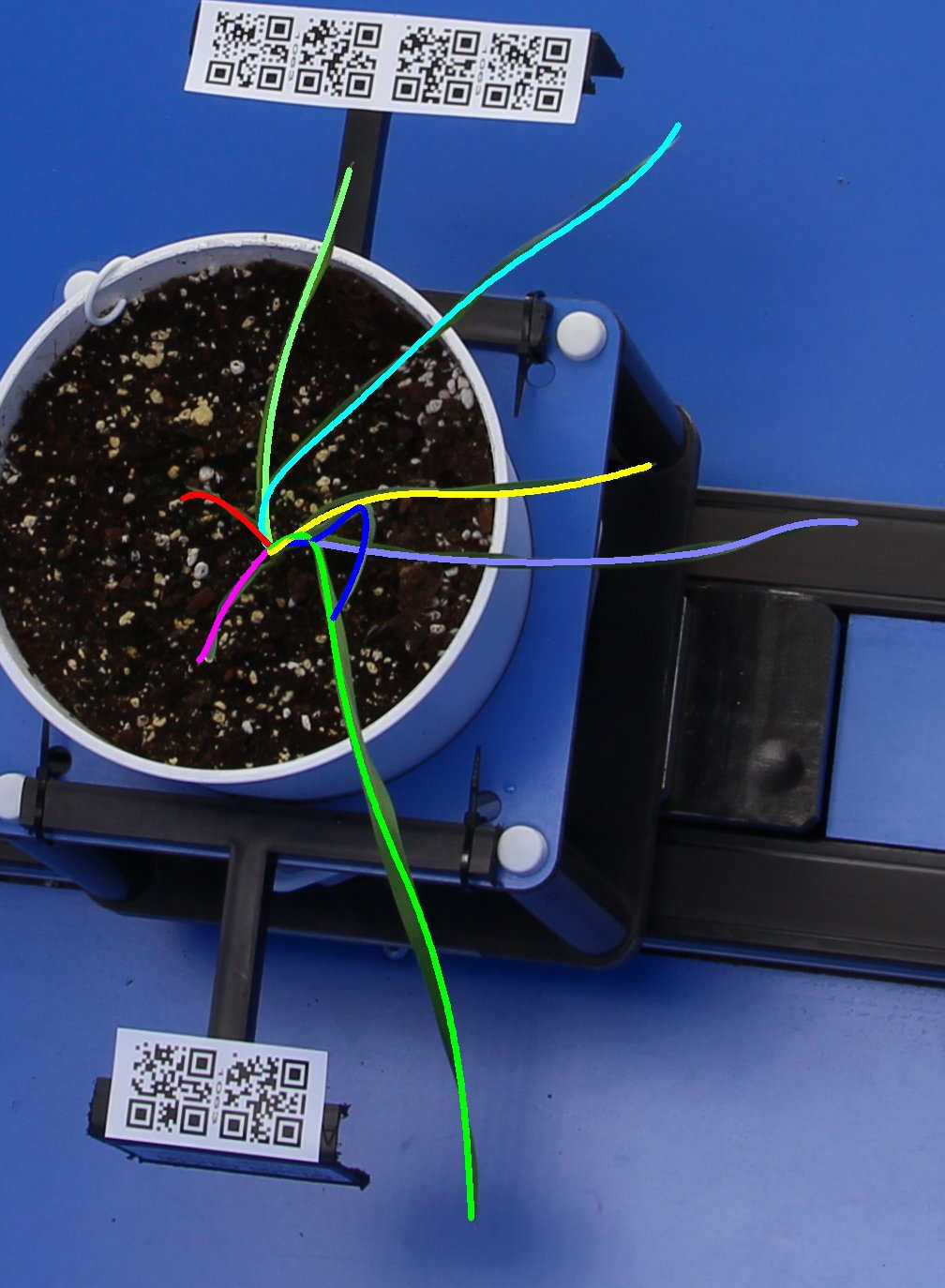}
}
\vspace{1mm}
\centerline{
\includegraphics[width=0.17\columnwidth]{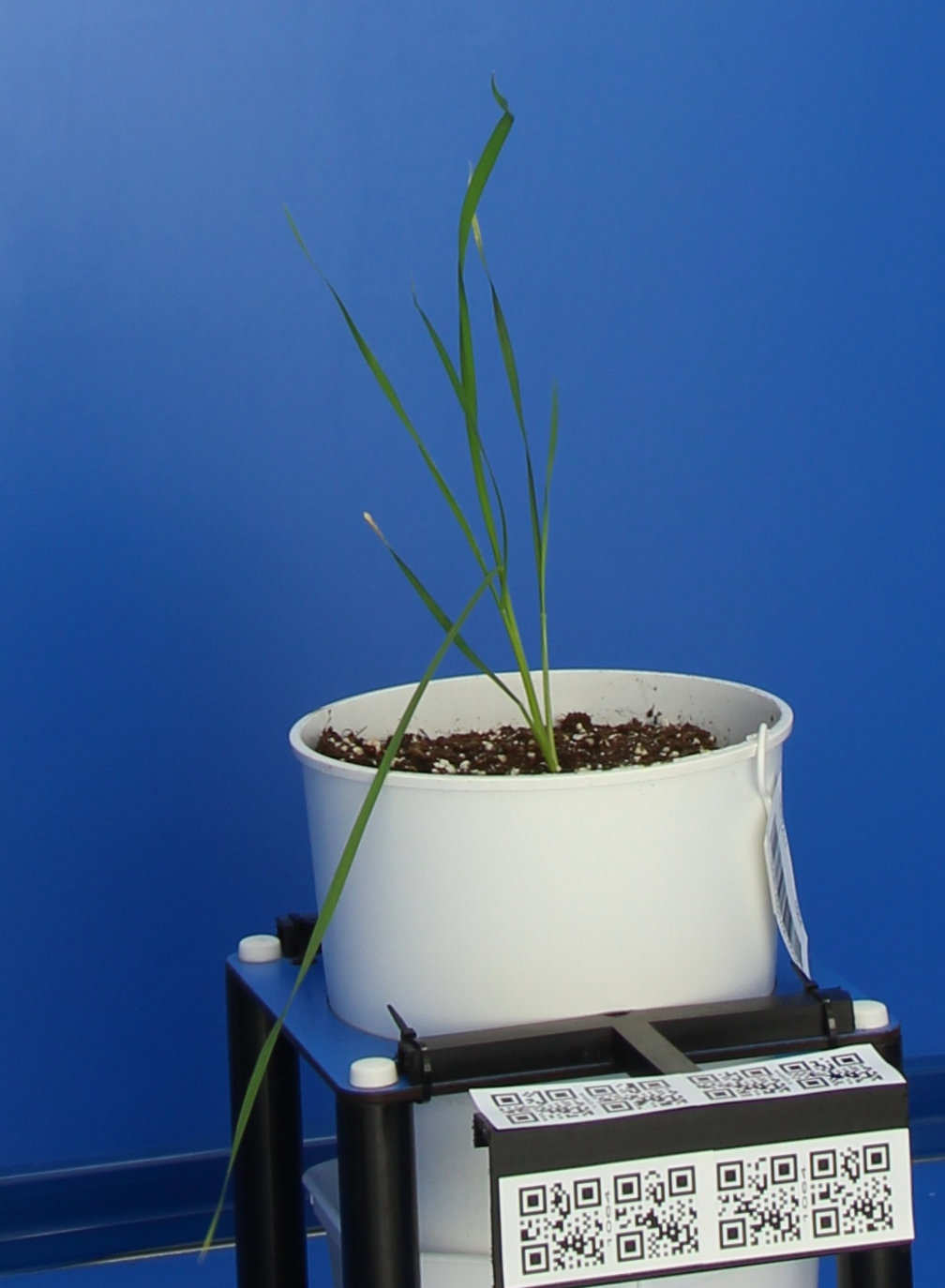}
\includegraphics[width=0.17\columnwidth]{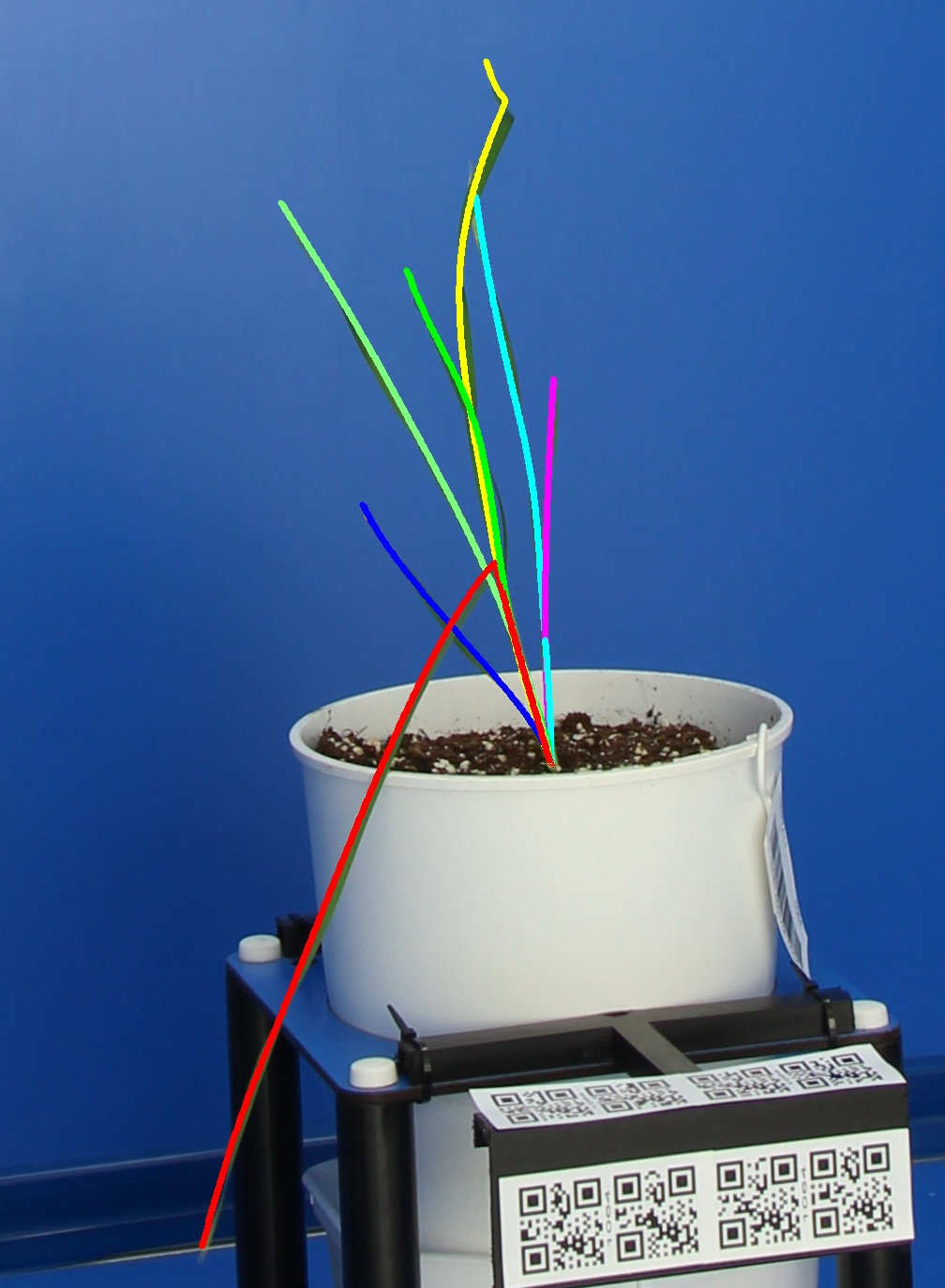}
\includegraphics[width=0.17\columnwidth]{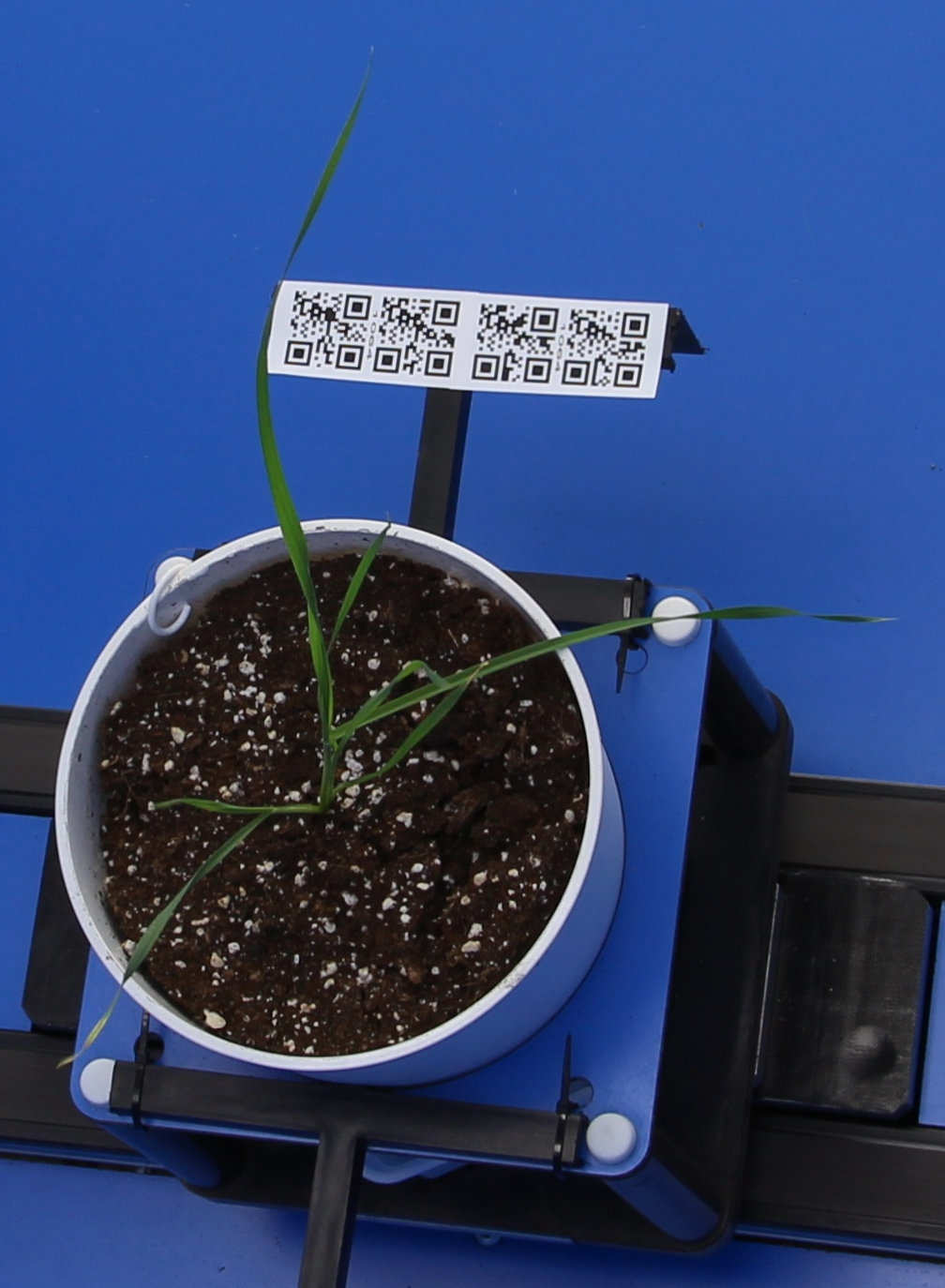}
\includegraphics[width=0.17\columnwidth]{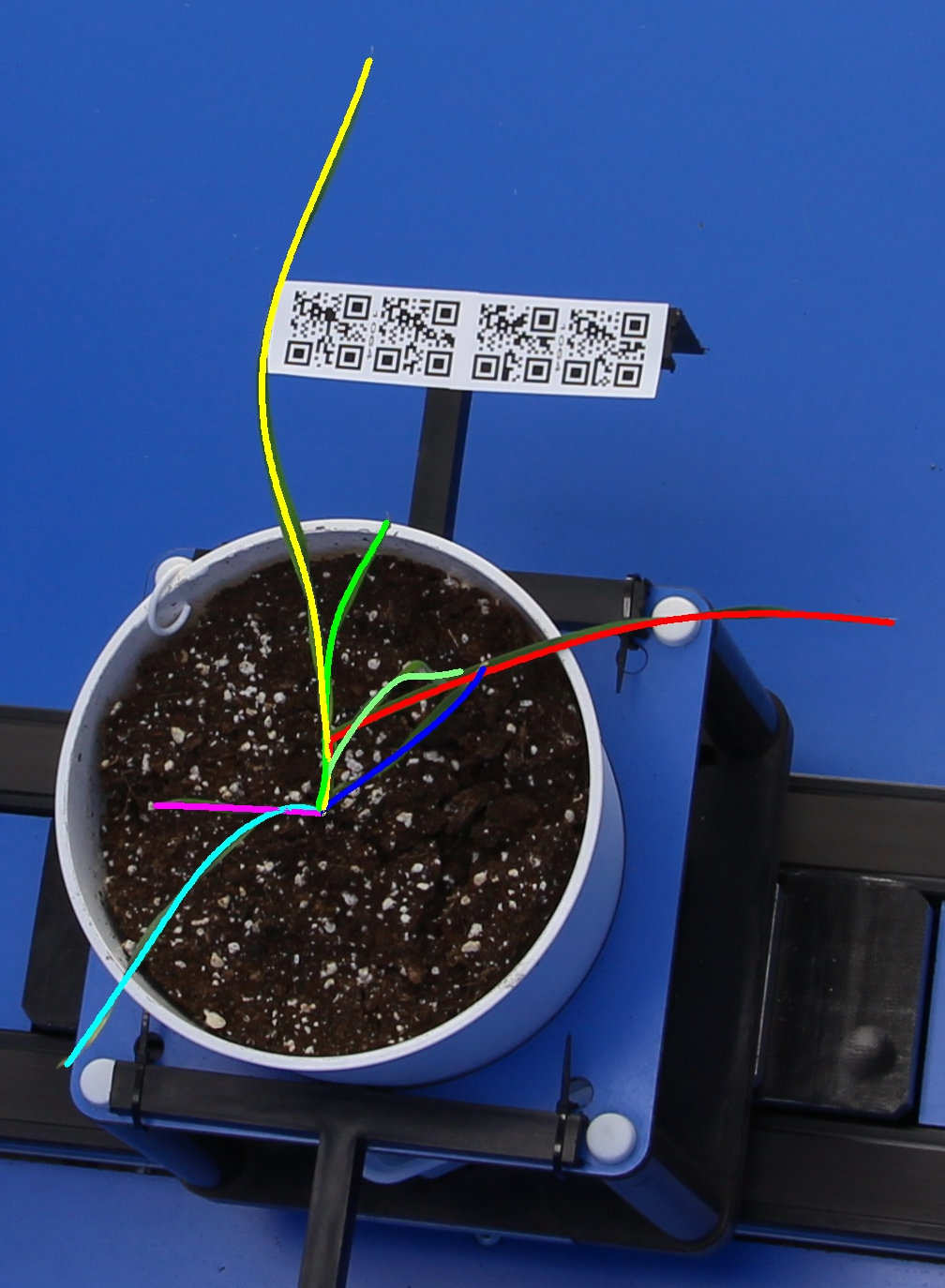}
}
\vspace{1mm}
\centerline{
\includegraphics[width=0.17\columnwidth]{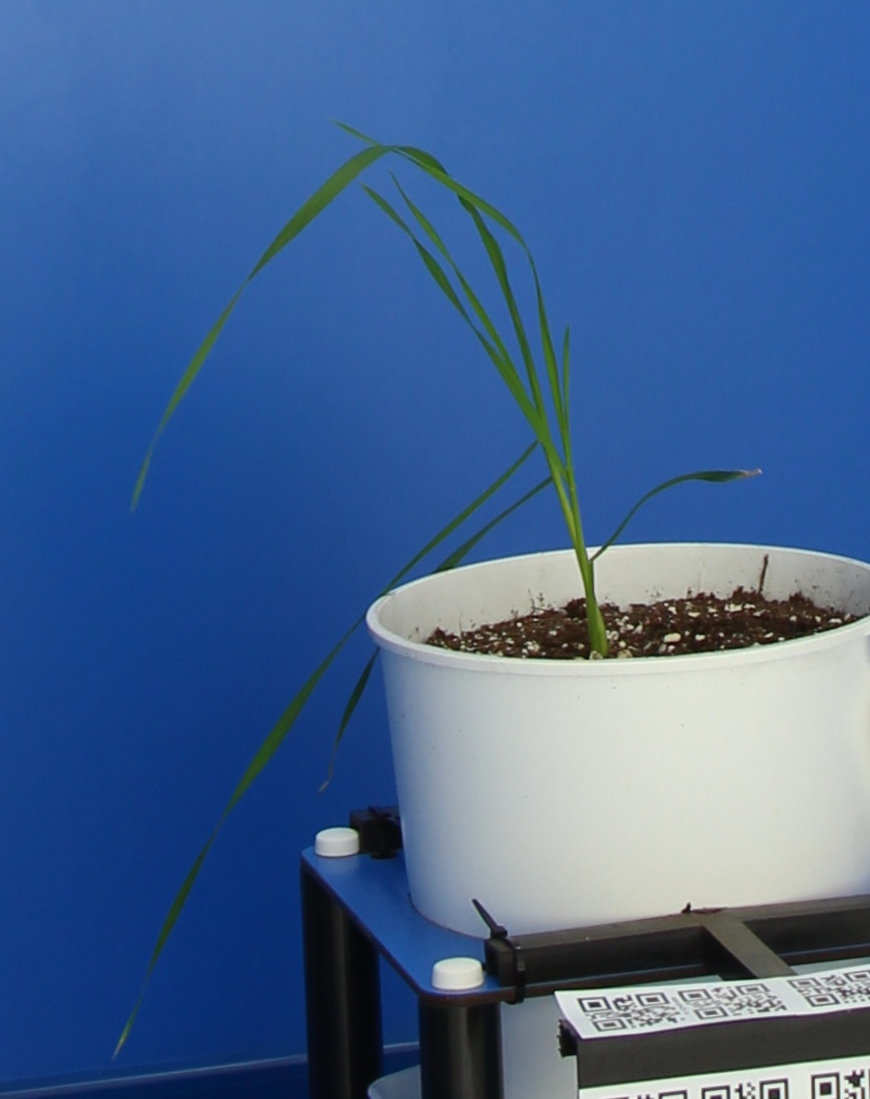}
\includegraphics[width=0.17\columnwidth]{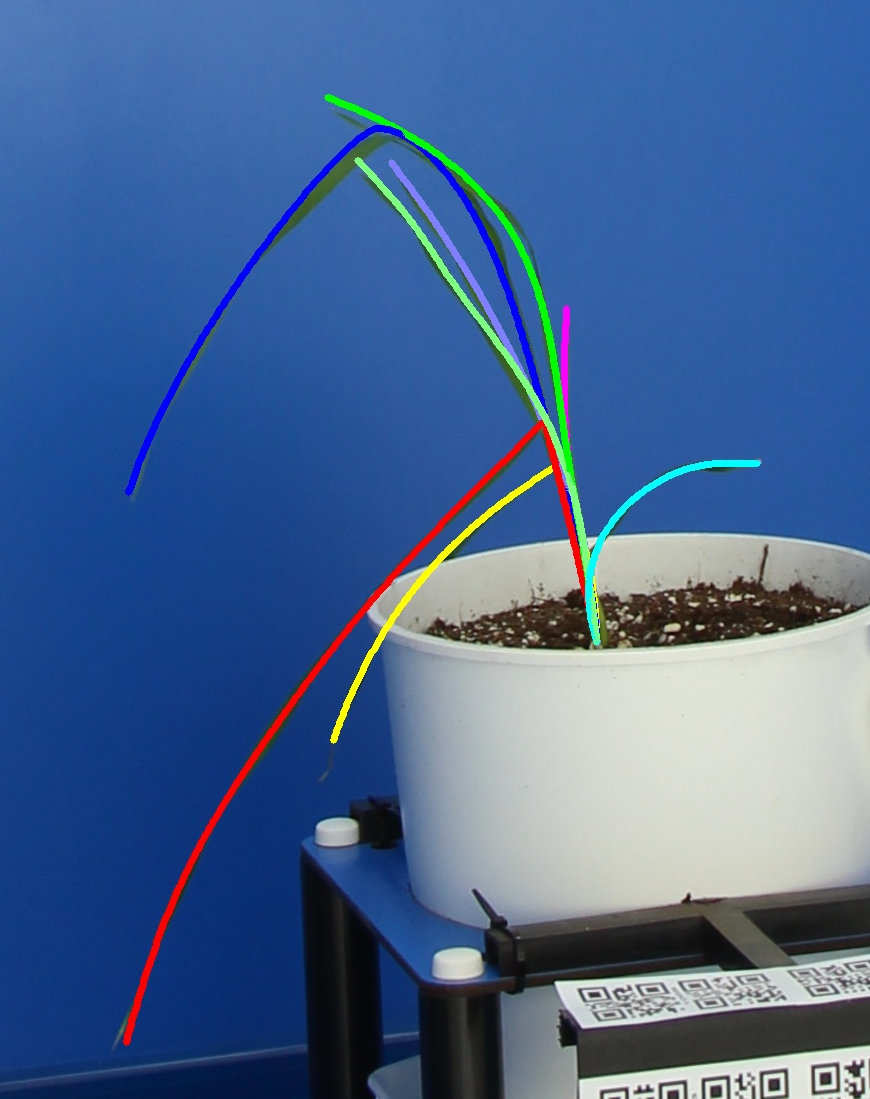}
\includegraphics[width=0.17\columnwidth]{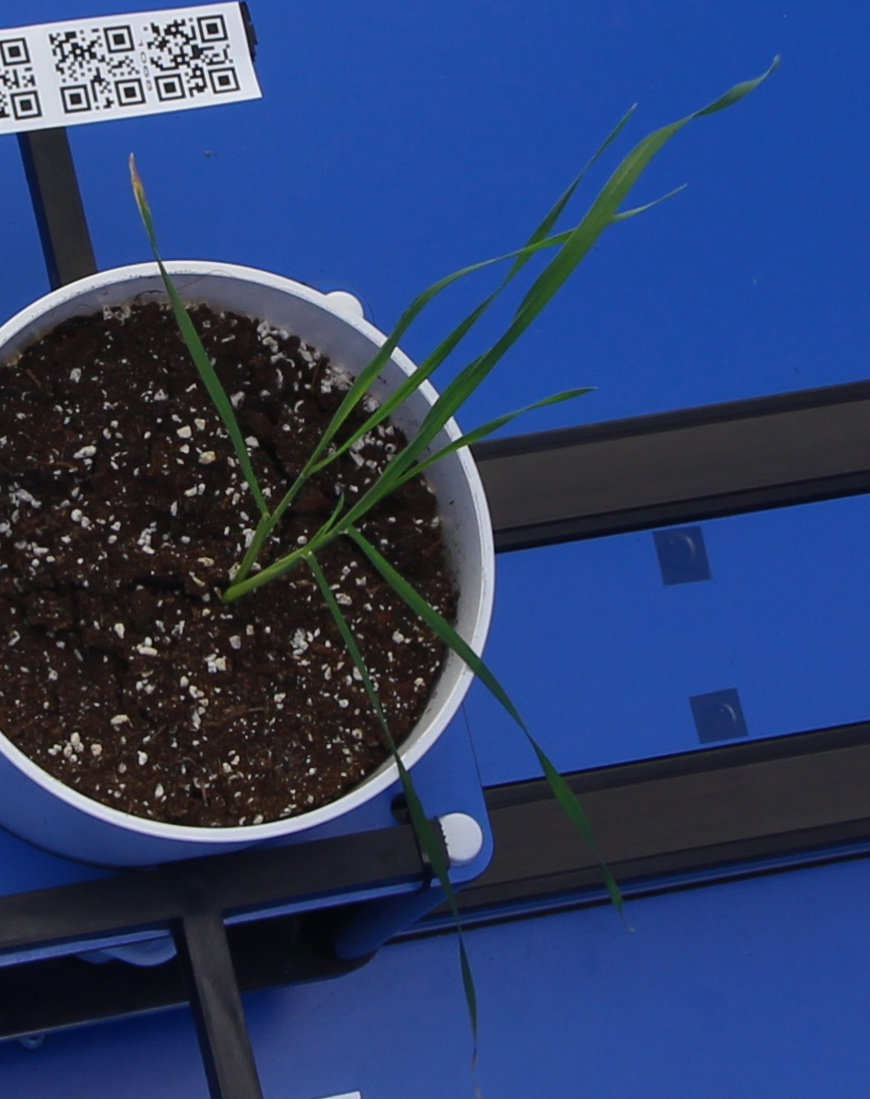}
\includegraphics[width=0.17\columnwidth]{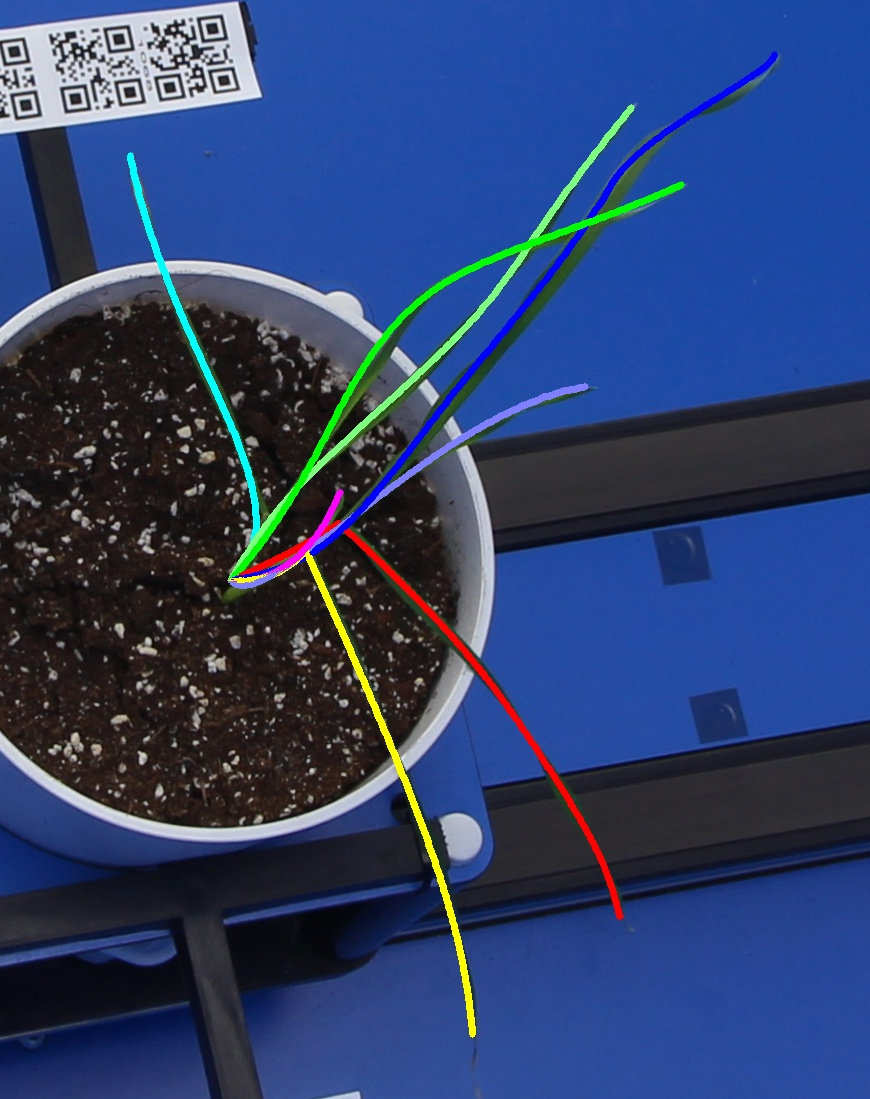}
}
\caption{Original images and reconstruction results}
\label{fig:res2}
\end{figure}

This method has been tested on a set of plants with up to $8$ leaves each, with manual measurements taken for the first $4$ leaves of each plant. Figure~\ref{fig:res2} shows the original images from two cameras, and the reconstructed plant model projected into those images, for $6$ plants. These results show the structure of the plant being accurately recovered despite overlap between multiple leaves. These results were generated with $50000$ runs of the model generation process, and with weights set to $\alpha=2 \times 10^{-7}$, $\beta=1.4$ and $\gamma=0.3$. Result images for more plants are included in Online Resource 1.

\begin{figure}[!tb]
\centerline{
\subfigure{\includegraphics[width=0.22\columnwidth]{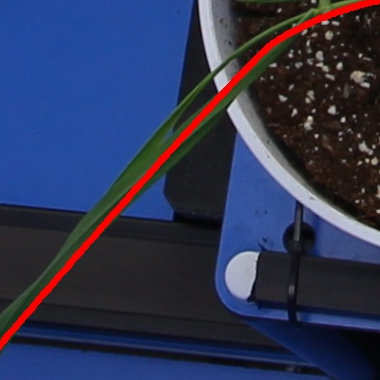}}
\subfigure{\includegraphics[width=0.22\columnwidth]{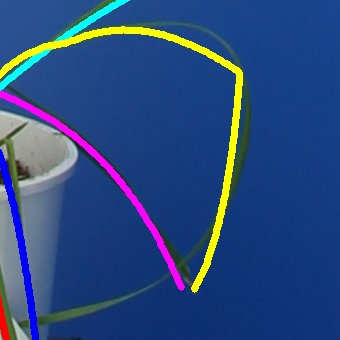}}
\subfigure{\includegraphics[width=0.22\columnwidth]{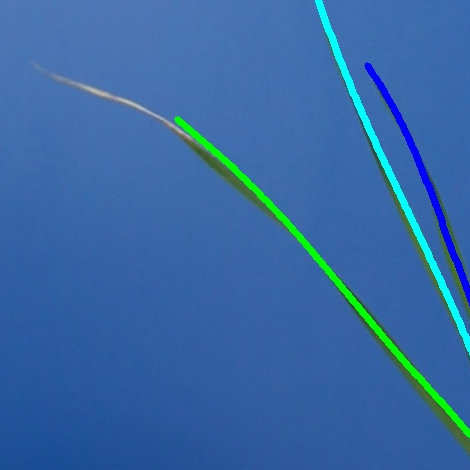}}
}
\caption{Partially reconstructed leaves}
\label{fig:partial}
\end{figure}

In most cases, the reconstruction process determined the correct number of leaves and generated a model close to the true shape of the plant. Figure~\ref{fig:partial} shows some cases where a leaf was not reconstructed, or was only partially reconstructed. In Figure~\ref{fig:partial}(a), a leaf was not reconstructed due to the leaf tip and most of the length of the leaf being occluded in all views by the leaf labelled in red. In Figure~\ref{fig:partial}(b), only part of the shape of the leaf labelled in yellow was recovered, as a close match for the leaf was not found in the database. This limitation would be improved with a more comprehensive model database. A leaf model was not fitted to the full extent of the leaf labelled in green in Figure~\ref{fig:partial}(c), due to the pixels of a dead leaf tip being classified as background during silhouette generation.

For this set of plants, we have compared leaf length measurements automatically extracted from the models with manual measurements of the first $4$ leaves of each plant. Manual measurements were taken from the leaf tip to the point at which the leaf meets the stem. To measure this distance from the reconstructed leaf models, we estimate this point by finding the point at which overlapping leaf models diverge. Table~\ref{tab:meas} shows automatically and manually measured leaf lengths in millimetres and relative percentage error for the set of plants seen in Figure~\ref{fig:res2}. For tests on a set of $40$ plants, the average difference between the manual measurements and our estimated leaf lengths was $19.06mm$. The average relative error was $8.64\%$.

This testing has highlighted an unforeseen ambiguity in the (stem-side) end point of such leaf measurements which leads to differences between the manually measured quantity and that estimated from the recovered structure.  It also indicates a need to conduct repeated manual measurements so as to estimate the error in that process.  Despite these limitations, the results show that the method is capable of automatically recovering meaningful plant structure estimates from image sets.

\begin{figure}[!tb]
\centerline{
\includegraphics[width=0.17\columnwidth]{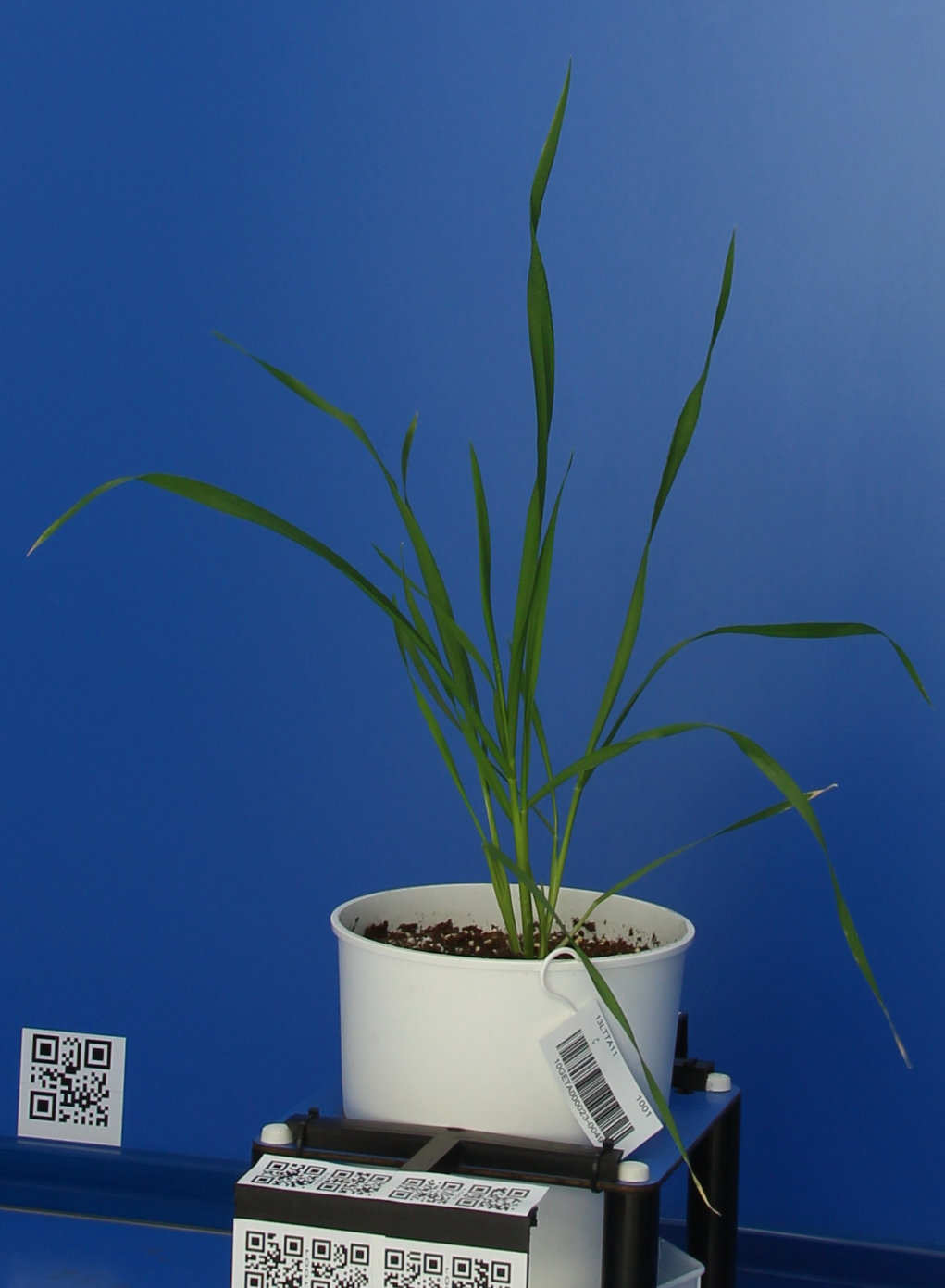}
\includegraphics[width=0.17\columnwidth]{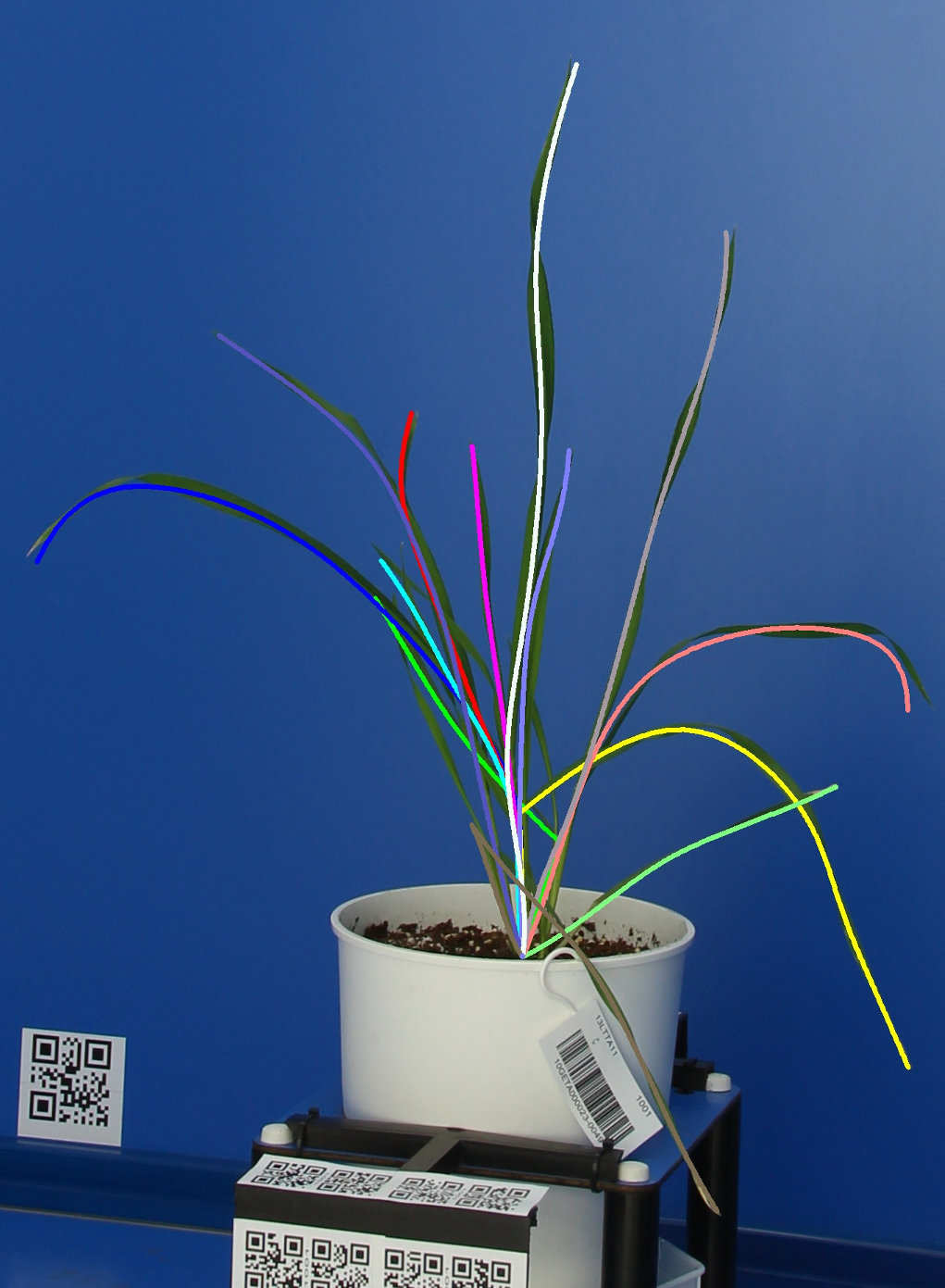}
\includegraphics[width=0.17\columnwidth]{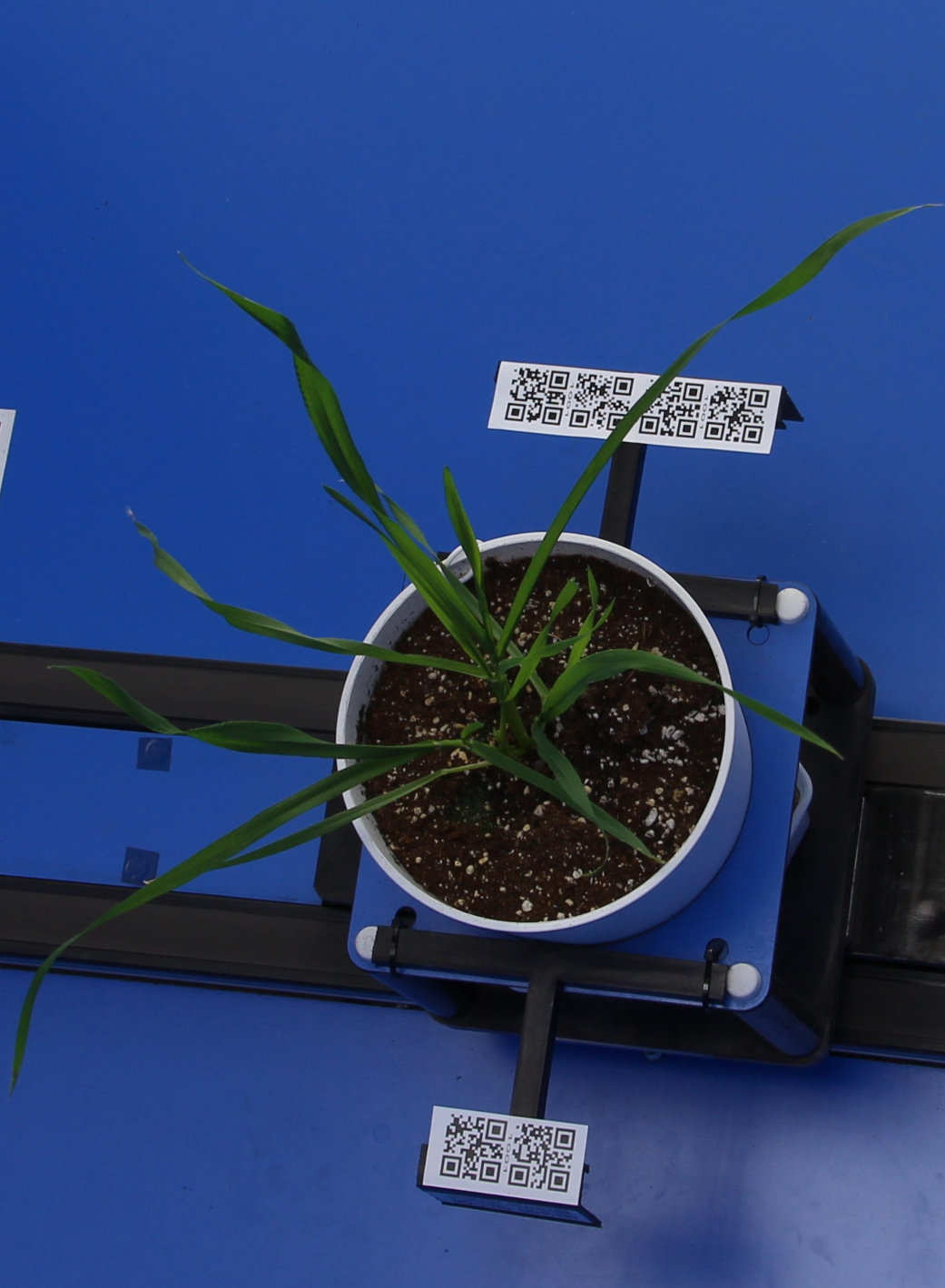}
\includegraphics[width=0.17\columnwidth]{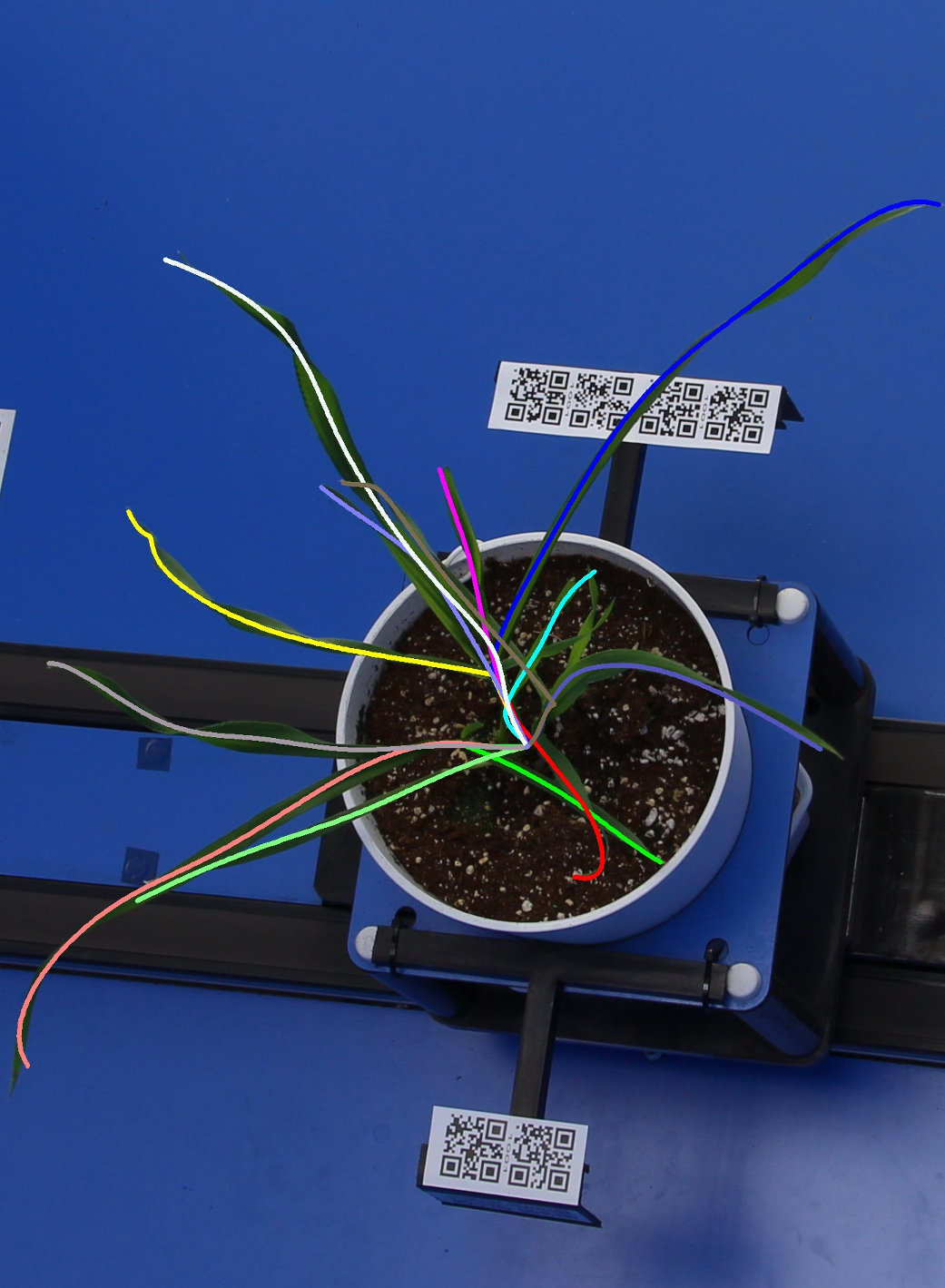}
}
\vspace{1mm}
\centerline{
\includegraphics[width=0.17\columnwidth]{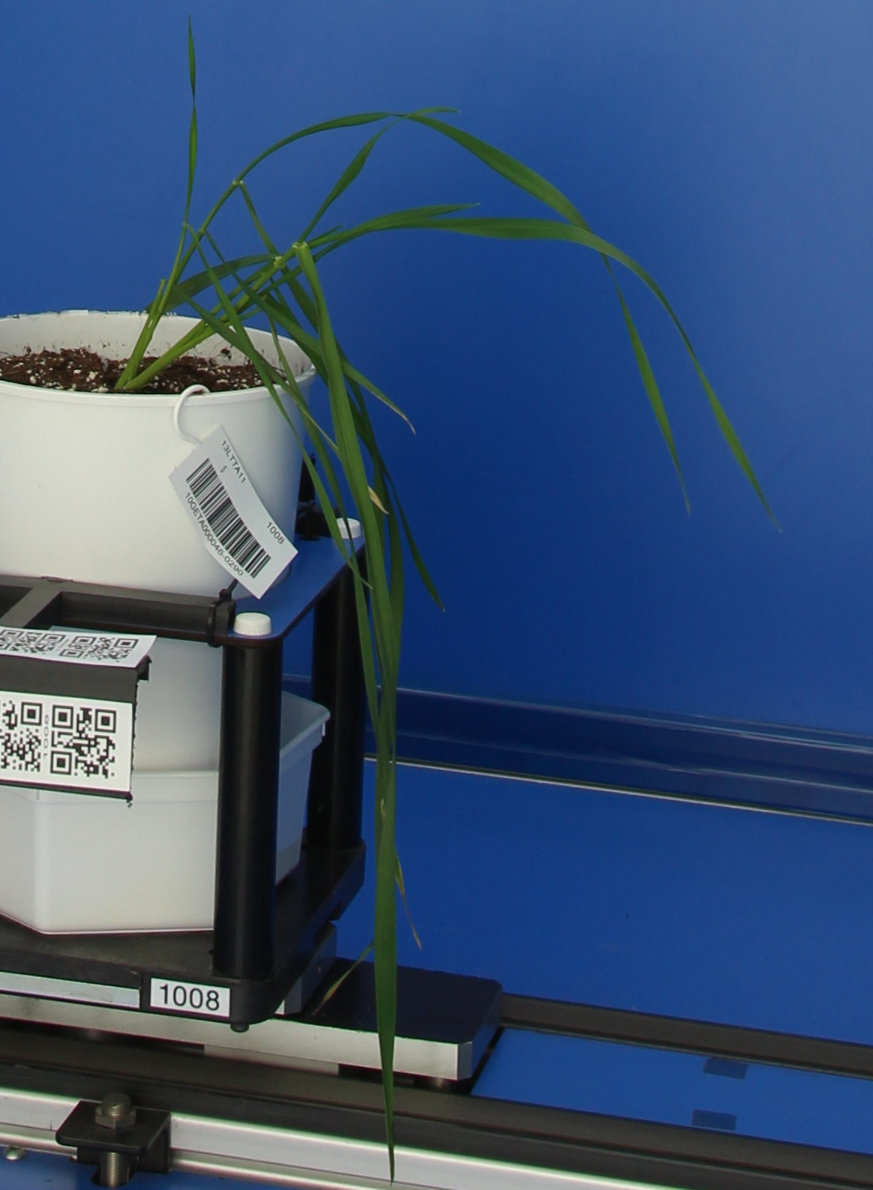}
\includegraphics[width=0.17\columnwidth]{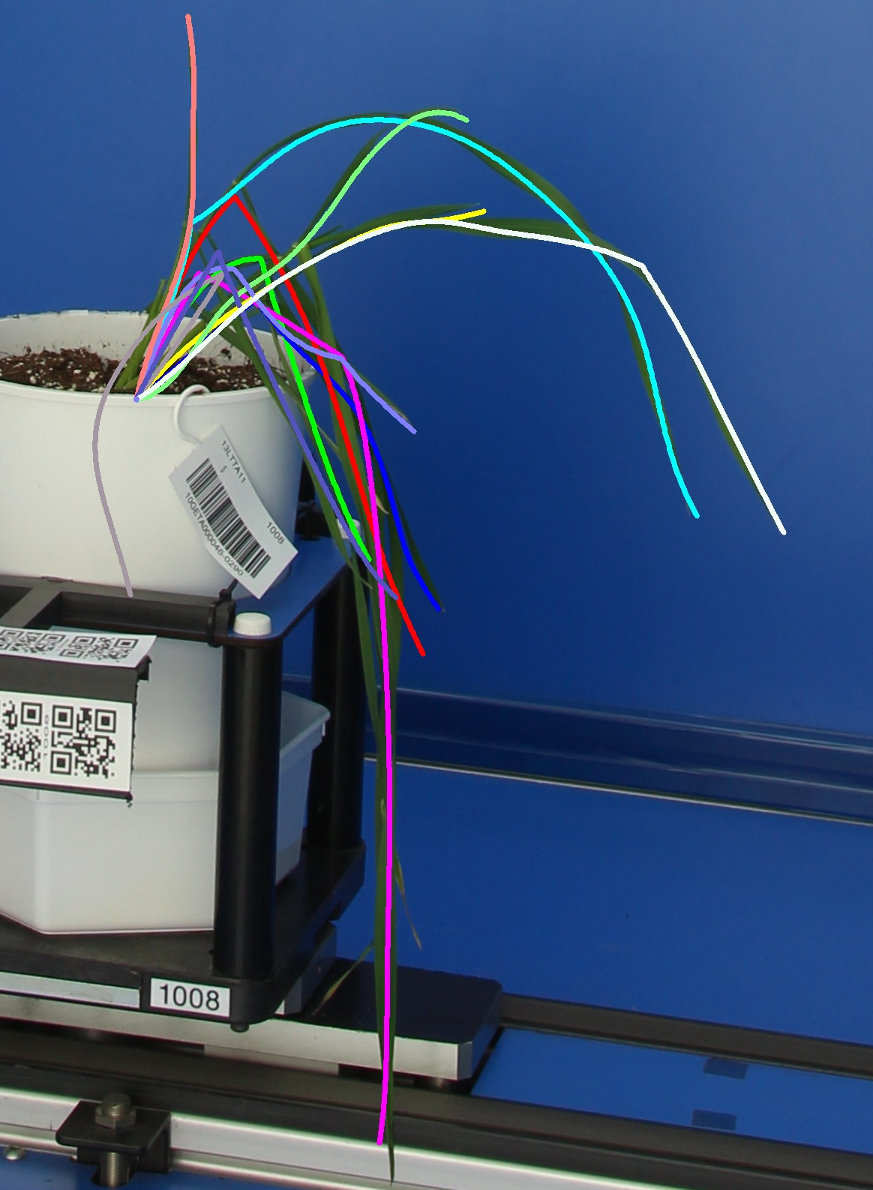}
\includegraphics[width=0.17\columnwidth]{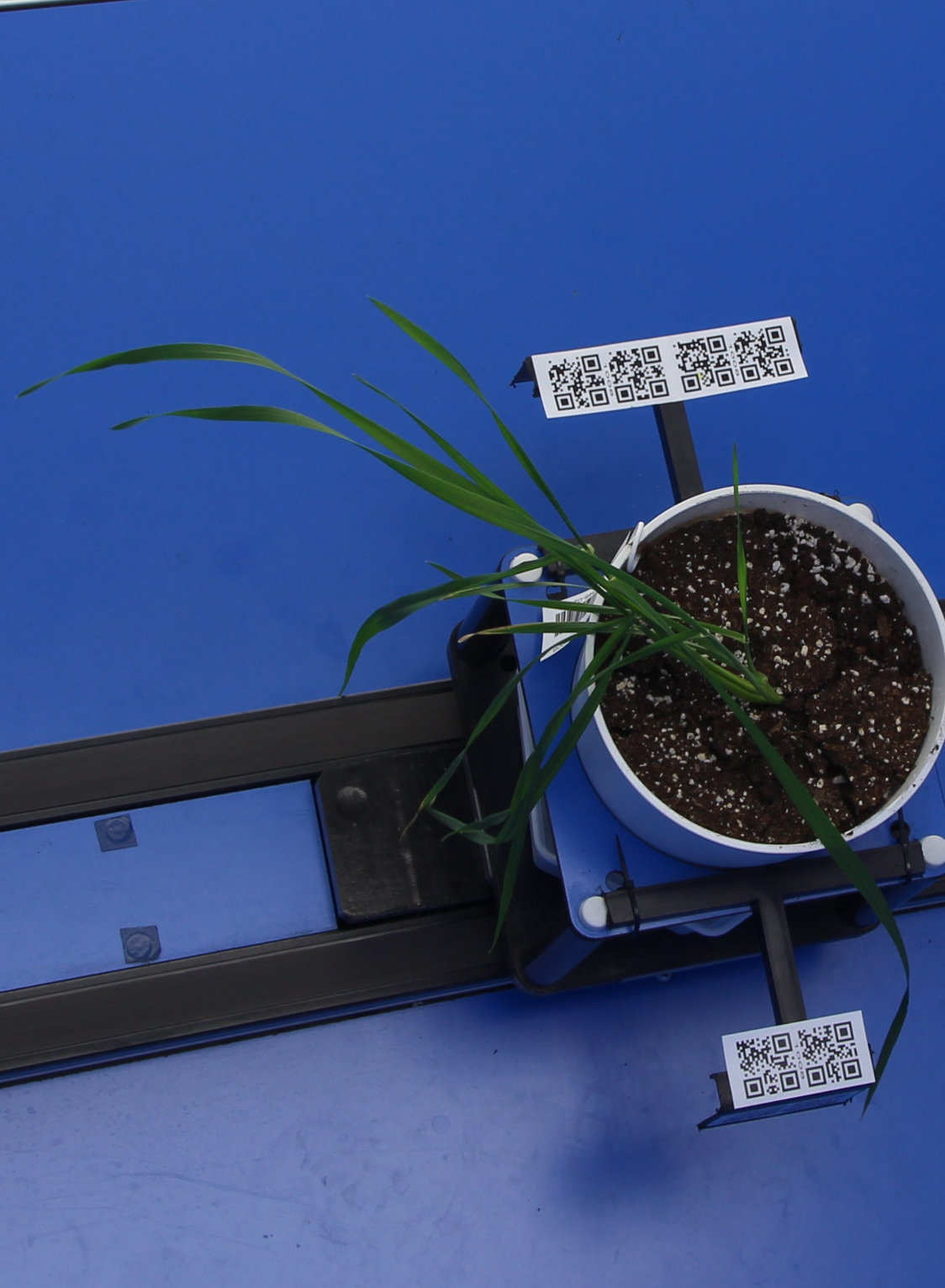}
\includegraphics[width=0.17\columnwidth]{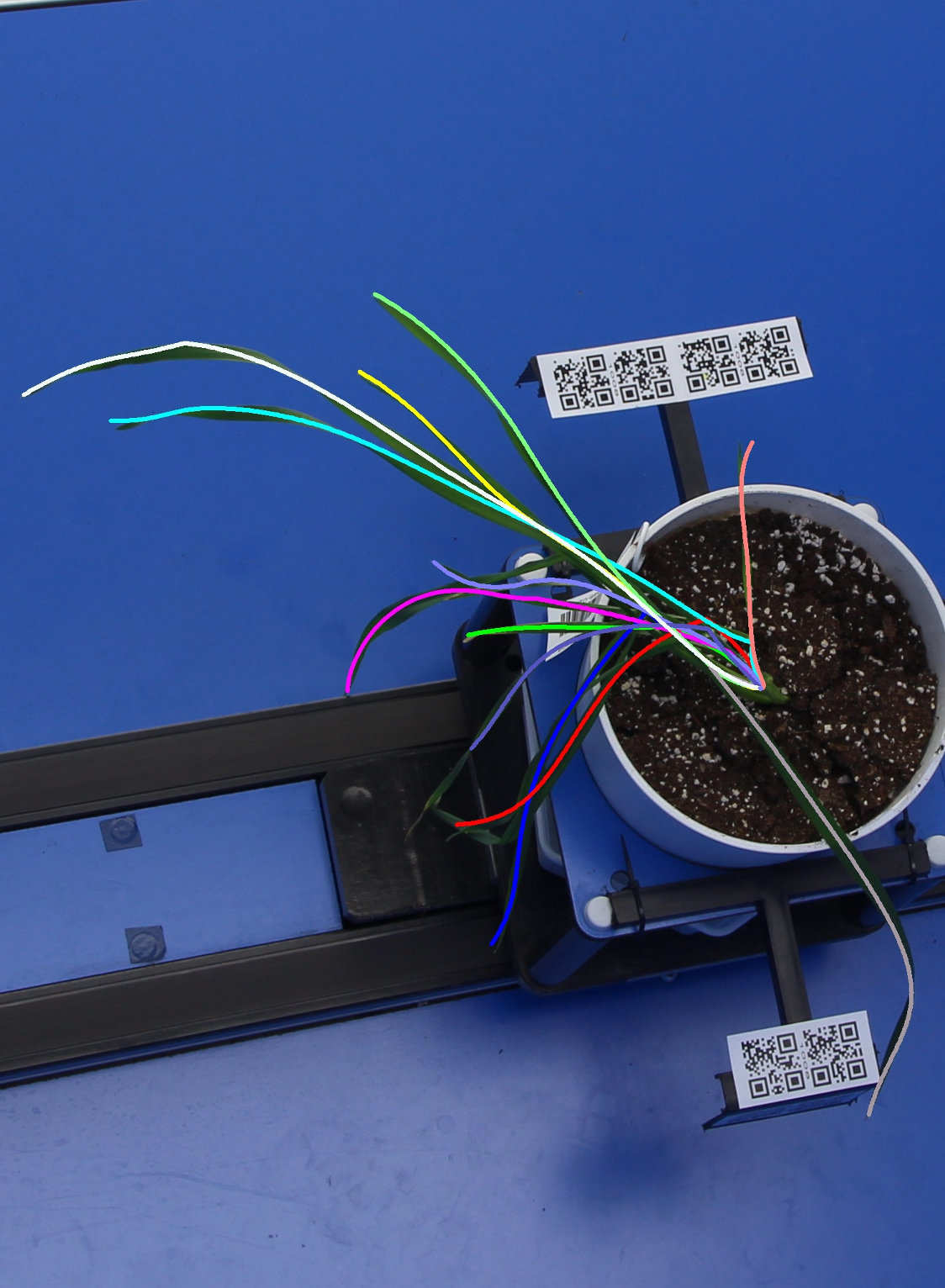}
}
\vspace{1mm}
\centerline{
\includegraphics[width=0.17\columnwidth]{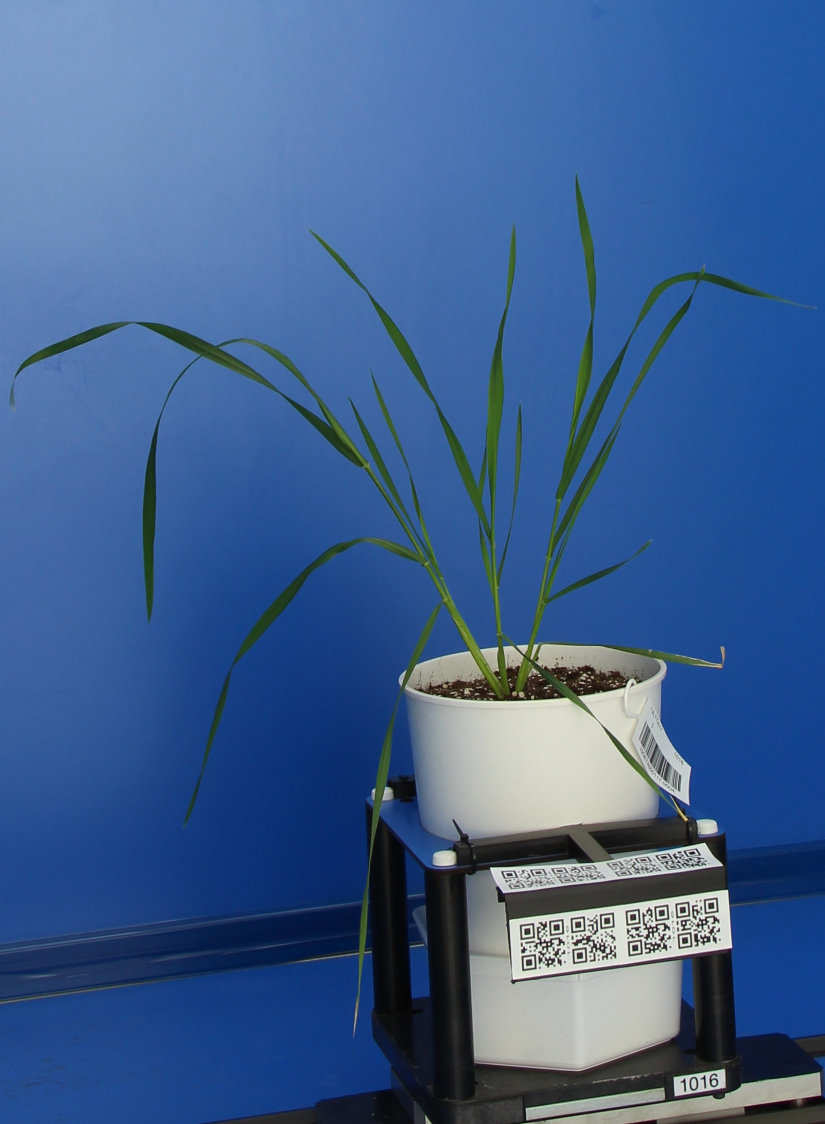}
\includegraphics[width=0.17\columnwidth]{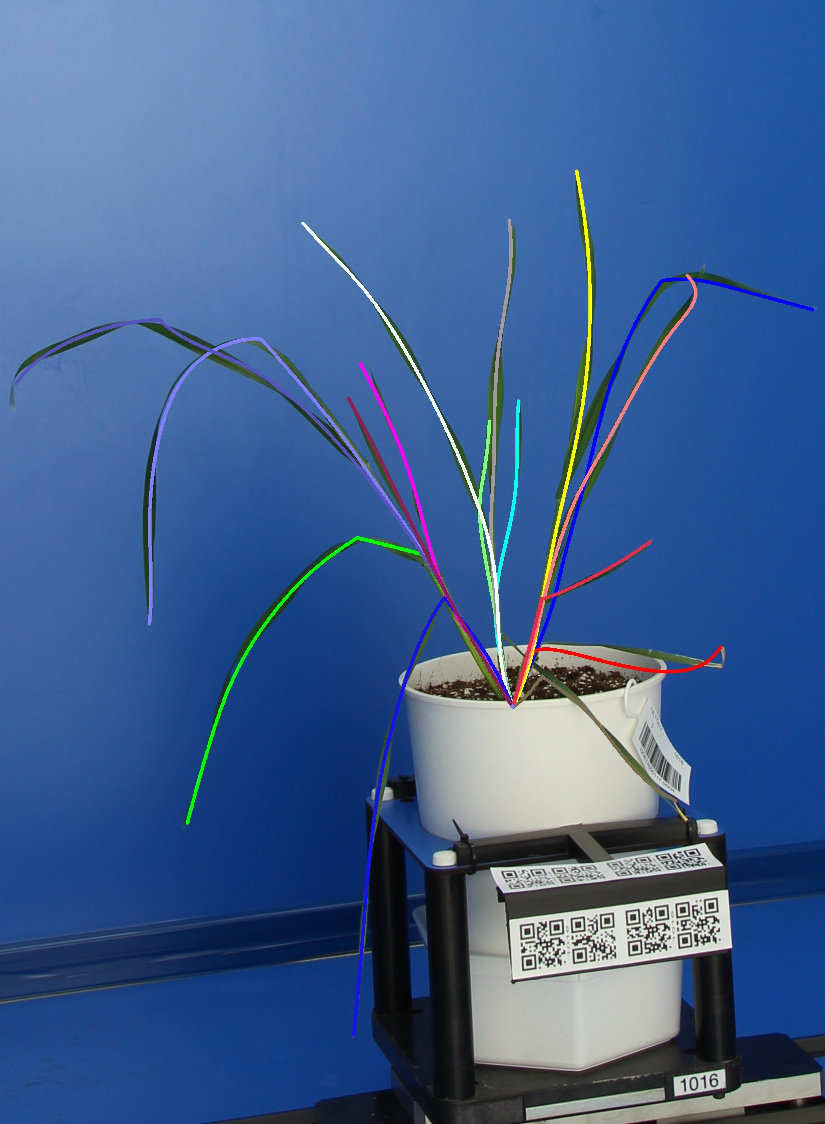}
\includegraphics[width=0.17\columnwidth]{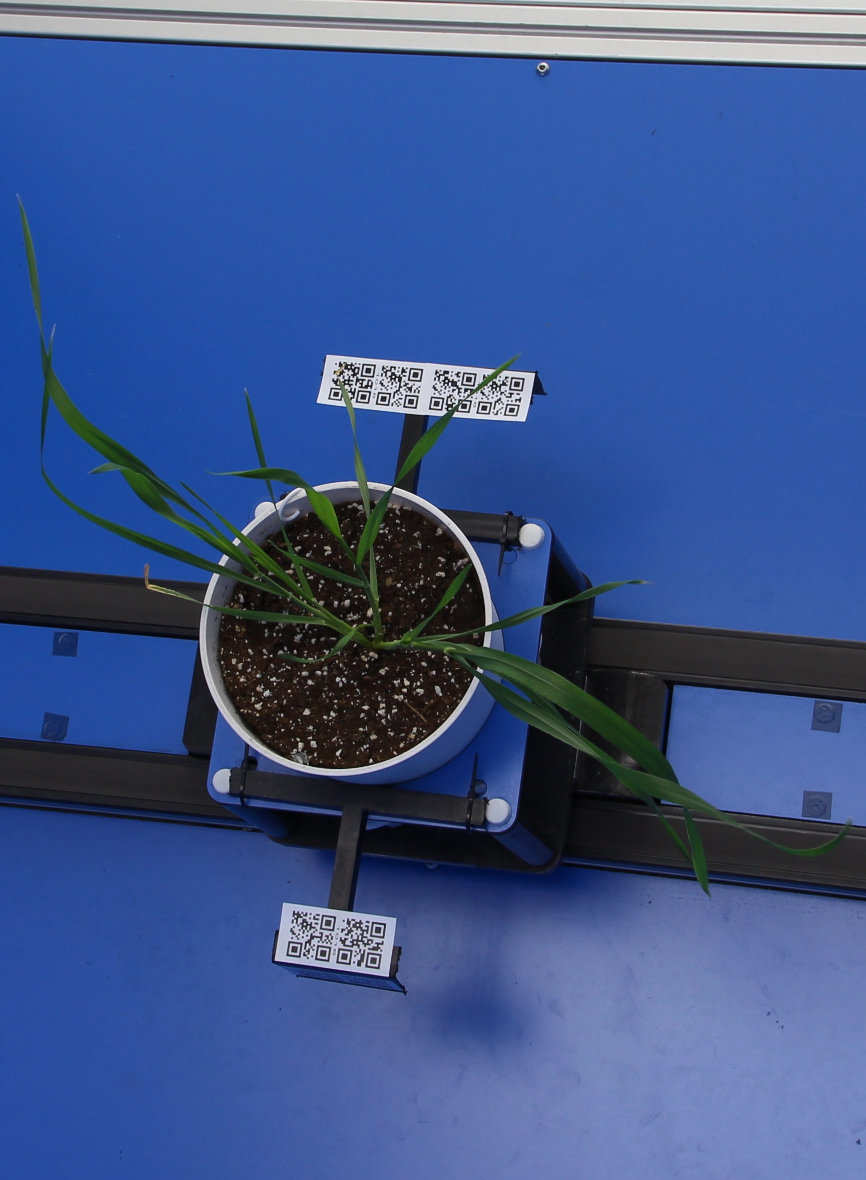}
\includegraphics[width=0.17\columnwidth]{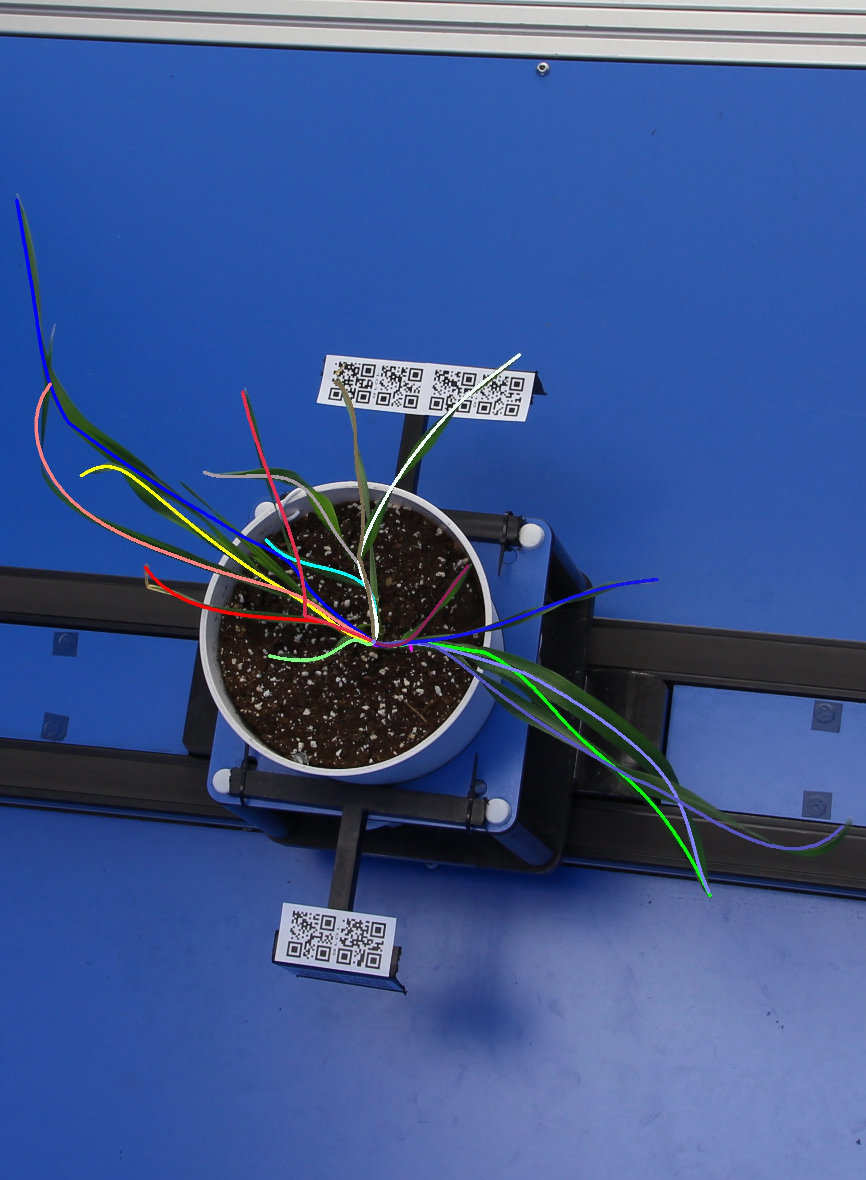}
}
\caption{Reconstruction results for more mature plants}
\label{fig:res_mat}
\end{figure}

Figure~\ref{fig:res_mat} shows results of applying this method to more mature plants with a greater density of leaves. In these cases, the structure of the majority of leaves was still recovered. However, some leaves with tips in regions where structure is dense were not identified, and the accuracy of the curves for the reconstructed leaves was also lower in these regions. Improving reconstruction accuracy for more mature plants will be a focus of further development of this method.

\begin{table}[th]
\caption{Measurement results for the first $4$ leaves}
\begin{center}\setlength{\tabcolsep}{5pt}
\begin{tabular}{l|cccc|cccc}
    \hline
    &\multicolumn{4}{c|}{\bfseries{Plant 1}} & \multicolumn{4}{c}{\bfseries{Plant 2}} \\
    Manual (mm) & 150.64 & 220.68 & 299.53 & 245.26 & 138.74 & 243.89 & 332 & 351 \\
    Estimated (mm) & 147.0 & 216.79 & 299.89 & 241.99 & 145.99 & 214.75 & 292.73 & 337.99  \\
    Relative (\%) & 2.42 & 1.77 & 1.55 & 0.92 & 5.23 & 11.95 & 11.83 & 3.71\\
    \hline
    \hline
    &\multicolumn{4}{c|}{\bfseries{Plant 3}} & \multicolumn{4}{c}{\bfseries{Plant 4}} \\
    Manual (mm) & 144.97 & 263.75 & 378 & 224.13 & 115.73 & 203.23 & 279.82 & 320.0 \\
    Estimated (mm)& 145.91 & 259.87 & 376.73 & 242.94 & 137.0 & 200.99 & 279.0 & 287.92 \\
    Relative (\%) & 0.65 & 1.47 & 0.34 & 8.39 & 17.51 & 1.1 & 0.29 & 10.02 \\
    \hline
    \hline
    &\multicolumn{4}{c|}{\bfseries{Plant 5}} & \multicolumn{4}{c}{\bfseries{Plant 6}} \\
    Manual (mm) & 101.4 & 185.82 & 259.16 & 299.87 & 119.22 & 211.86 & 273.85 & 304.55 \\
    Estimated (mm) & 117.0 & 162.0 & 255.81 & 251.98 & 130.99 & 184.51 & 272.54 & 265.98 \\
    Relative  (\%) & 15.38 & 12.82 & 1.29 & 15.97 & 9.87 & 12.91 & 0.48 & 12.66 \\
    \hline
\end{tabular}
\end{center}
\label{tab:meas}
\end{table}

\section{Conclusions and Future Work}

We have presented a method suitable for recovering the structure of thin plants from a small set of images captured by widely spaced cameras. There are a range of potential future developments for this method. Although the present method operates only on RGB images, it would be straightforward to incorporate depth map information into the fitting process, allowing for reconstruction using depth camera or laser data from a limited range of views.

The method could potentially be applied to single images, using the variation in plausible reconstructions of the image to determine the range of possible values for various plant properties. The method could also provide a means of estimating further physical properties of  leaves from measured properties of leaves represented in the database. The structure estimates can be used for leaf angle and length measurements, and we plan to use these paths as a basis for also measuring leaf width and senescence.

We plan to use the estimated structures of plants over time to track plant growth, with a database of models of developing plants used to determine plausible matches between the estimated leaves at different time steps. The method will also be refined to improve the reconstruction accuracy for more mature plants, where the structure of individual leaves is more difficult to distinguish using only skeletons extracted from silhouettes.

\clearpage

\bibliographystyle{splncs03}
\bibliography{main}
\end{document}